\documentclass{article} % For LaTeX2e
\usepackage{iclr2025_conference,times}

\usepackage{amsmath,amsfonts,bm}

\def\eqref#1{equation~\ref{#1}}
\def\1{\bm{1}}

\DeclareMathAlphabet{\mathsfit}{\encodingdefault}{\sfdefault}{m}{sl}
\SetMathAlphabet{\mathsfit}{bold}{\encodingdefault}{\sfdefault}{bx}{n}

\DeclareMathOperator{\Tr}{Tr}

\usepackage{hyperref}
\usepackage[ruled,vlined]{algorithm2e}
\definecolor{commentcolor}{RGB}{110,154,155} 
\newcommand{\PyComment}[1]{\ttfamily\textcolor{commentcolor}{\# #1}}  % add a "#" before the input text "#1"
\newcommand{\PyCode}[1]{\ttfamily\textcolor{black}{#1}} % \ttfamily 
\newcommand{\PyCodeHighlight}[1]{\ttfamily\textcolor{red}{#1}} % \ttfamily is the code font

\usepackage{etoolbox}
\usepackage{url}
\usepackage{listings}
\usepackage{enumitem}

\makeatletter
\patchcmd{\@algocf@start}% <cmd>
  {-1.5em}% <search>
  {0pt}% <replace>
  {}{}% <success><failure>
\makeatother

\SetAlFnt{\small}
\SetAlCapFnt{\small}
\SetAlCapNameFnt{\small}
\SetAlCapHSkip{0pt}

\usepackage{xspace}
\usepackage{cleveref}
\usepackage{subcaption}
\usepackage{xcolor,colortbl}
\usepackage{makecell}
\usepackage{array}
\usepackage[export]{adjustbox}
\usepackage{tabularray}
\UseTblrLibrary{booktabs}
\makeatletter
\DeclareRobustCommand\onedot{\futurelet\@let@token\@onedot}
\def\@onedot{\ifx\@let@token.\else.\null\fi\xspace}

\def\eg{\emph{e.g}\onedot} 
\def\ie{\emph{i.e}\onedot}

\makeatother

\newcommand\rot[1]{\rotatebox{90}{#1}}
\renewcommand\Tr[1]{\begin{tabular}[t]{@{}c@{}}#1\end{tabular}}

\title{ImmersePro: End-to-End Stereo Video Synthesis Via Implicit Disparity Learning}

\author{Jian Shi, Zhenyu Li, Peter Wonka 
\\
KAUST\\
\texttt{\{jian.shi,zhenyu.li.1,peter.wonka\}@kaust.edu.sa} \\
}

\iclrfinalcopy % Uncomment for camera-ready version, but NOT for submission.
\begin{document}

\maketitle

\begin{abstract}
We introduce \textit{ImmersePro}, an innovative framework specifically designed to transform single-view videos into stereo videos. This framework utilizes a novel dual-branch architecture comprising a disparity branch and a context branch on video data by leveraging spatial-temporal attention mechanisms. \textit{ImmersePro} employs implicit disparity guidance, enabling the generation of stereo pairs from video sequences without the need for explicit disparity maps, thus reducing potential errors associated with disparity estimation models.
In addition to the technical advancements, we introduce the YouTube-SBS dataset, a comprehensive collection of 423 stereo videos sourced from YouTube. This dataset is unprecedented in its scale, featuring over 7 million stereo pairs, and is designed to facilitate training and benchmarking of stereo video generation models. Our experiments demonstrate the effectiveness of \textit{ImmersePro} in producing high-quality stereo videos, offering significant improvements over existing methods.
Compared to the best competitor stereo-from-mono we quantitatively improve the results by 11.76\% (L1), 6.39\% (SSIM), and 5.10\% (PSNR).
\end{abstract}

\section{Introduction}

A stereo movie, also known as a 3D movie, provides three-dimensional visual effects by employing stereoscopic techniques. By capturing or creating separate views for the left and right eyes, a 3D immersive experience can be achieved by using dedicated hardware such as head-mounted displays or autostereoscopic displays.
The disparity between the two views perceived by the viewer's brain creates the illusion of depth, making the objects in the movie appear at varying distances, thereby enhancing the immersive experience of the film.
Shooting stereo movies in the film industry often involves high costs due to the need for specialized equipment and meticulous post-production processes.
Alternatively, the stereoscopic effect can be created through a post-production process for videos that are shot with monocular cameras.
This post-production process uses \textit{stereo conversion}, which adds the binocular disparity depth cue to digital images.
It requires significant manual work by artists since inaccurate depth mapping and misrepresentations of occluded areas can cause visual discomfort~\cite{Devernay2010}.
% This \textit{stereo conversion} strategy is naturally more generalized (\ie can convert movies captured with monocular cameras) and cheaper.
In this paper, we propose an automated system that can reduce the time and expense associated with the conversion process, making it more accessible and economically feasible for more films.

% To enhance the viewing experience, films might employ a stronger stereoscopic effect at the start and end, while moderating it in the middle to ensure viewer comfort~\cite{neuman2009bolt,Ranftl2022}.
% Automated systems can maintain consistent quality throughout the conversion process, reducing the likelihood of human error and variability in the depth perception across different scenes.
% The stereo parameters of stereo videos are hard to retrieve. Compared to other stereo videos (\eg stereo advertisement with strong stereoscopic effect), stereo movie videos contain more consistent stereoscopic effects for prolonged viewing.
% To eleminate the

\begin{figure}[h]
     \centering
     \begin{subfigure}[b]{0.245\textwidth}
         \centering
         \includegraphics[width=\textwidth,height=.9\textwidth]
         {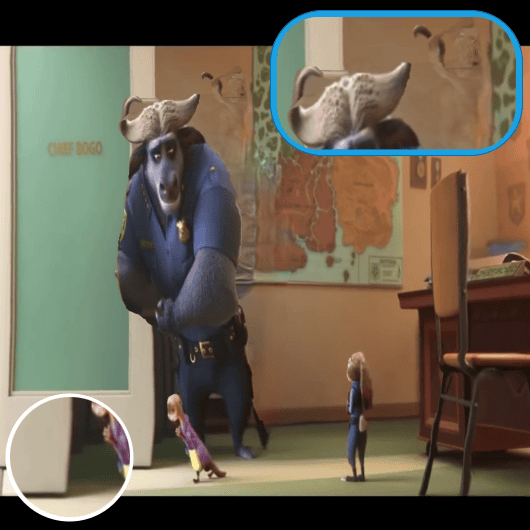}
         \includegraphics[width=\textwidth,height=.9\textwidth]
         {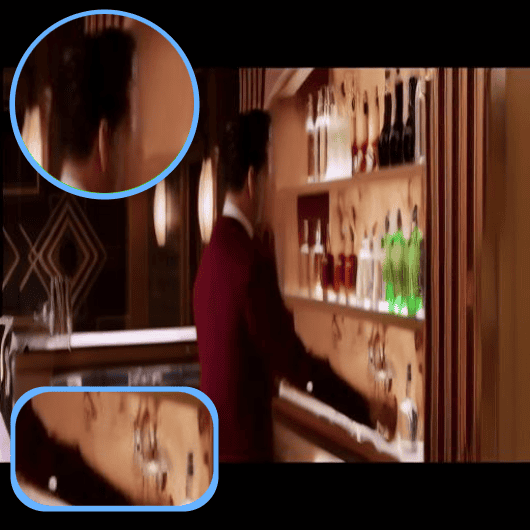}
         \includegraphics[width=\textwidth,height=.9\textwidth]
         {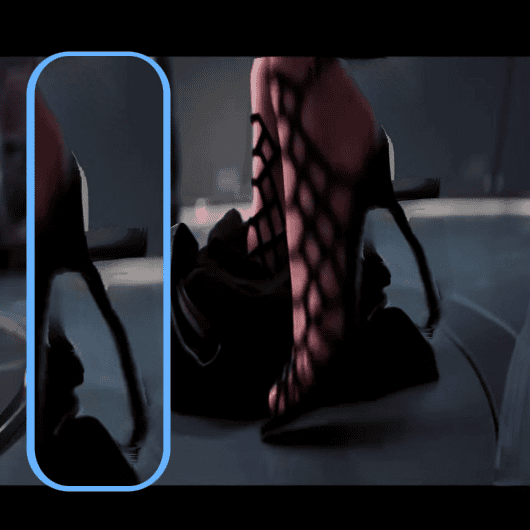}
         \caption{3D Photo}
     \end{subfigure}%
     \begin{subfigure}[b]{0.245\textwidth}
         \centering
         \includegraphics[width=\textwidth,height=.9\textwidth]
         {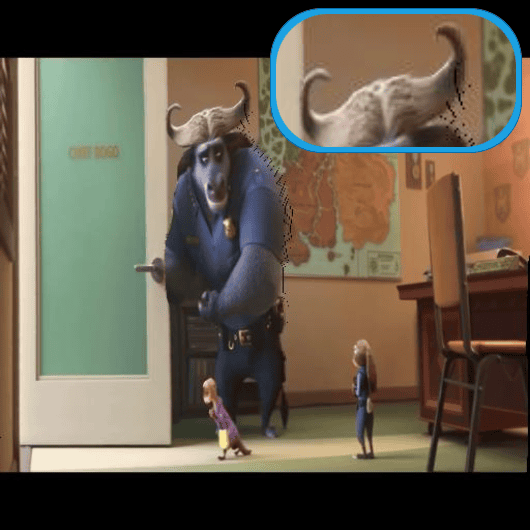}
         \includegraphics[width=\textwidth,height=.9\textwidth]
         {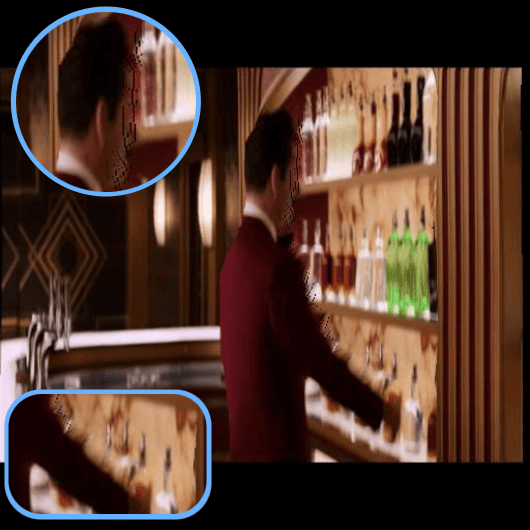}
         \includegraphics[width=\textwidth,height=.9\textwidth]
         {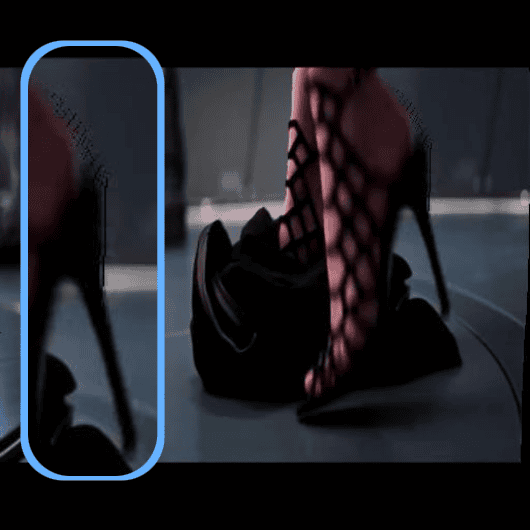}
         \caption{Stereo From Mono}
     \end{subfigure}%
     \begin{subfigure}[b]{0.245\textwidth}
         \centering
         \includegraphics[width=\textwidth,height=.9\textwidth]
         {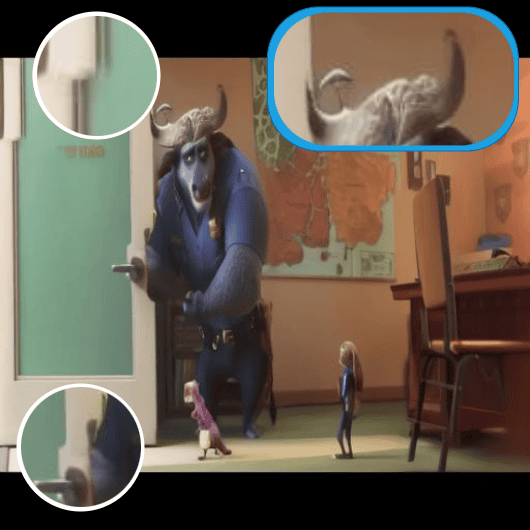}
         \includegraphics[width=\textwidth,height=.9\textwidth]
         {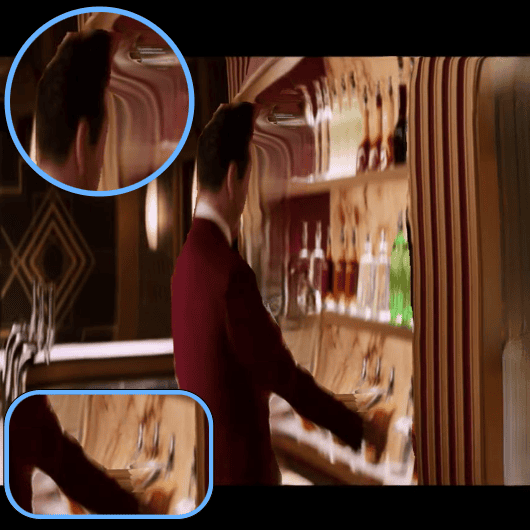}
         \includegraphics[width=\textwidth,height=.9\textwidth]
         {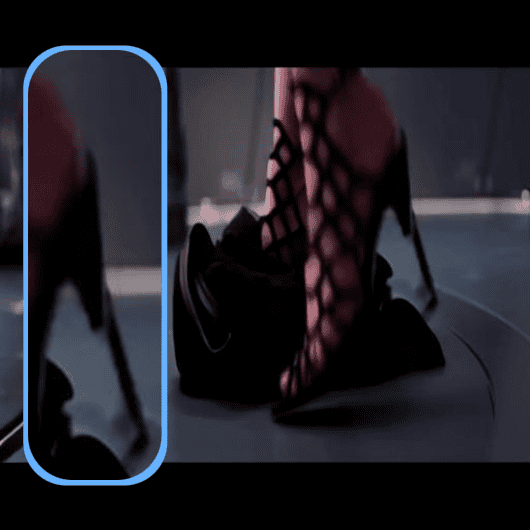}
         \caption{Stereo Diffusion}
     \end{subfigure}%
     \begin{subfigure}[b]{0.245\textwidth}
         \centering
         \includegraphics[width=\textwidth,height=.9\textwidth]
         {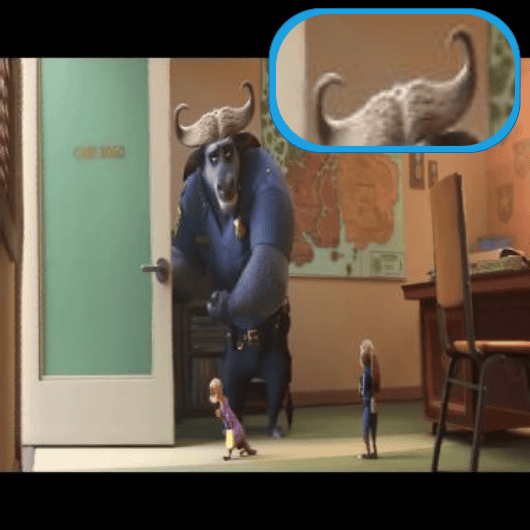}
         \includegraphics[width=\textwidth,height=.9\textwidth]
         {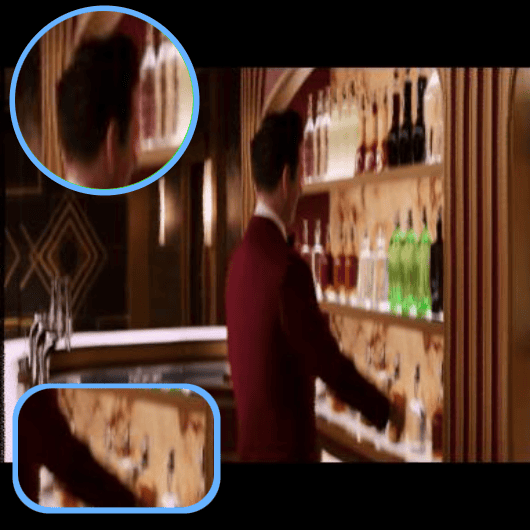}
         \includegraphics[width=\textwidth,height=.9\textwidth]
         {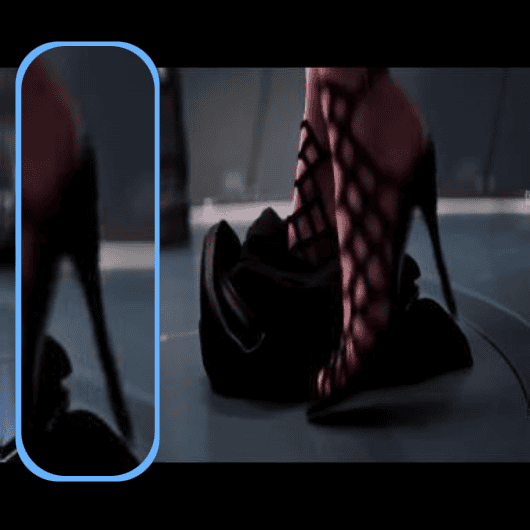}
         \caption{Ours}
     \end{subfigure}
  \caption{ImmersePro is a video method to convert a single-view video to a stereo video by predicting plausible right-view images for each input frame. Compared to previous work processing images frame by frame (3D Photo or Stereo from Mono), our method has the best visual quality.}
  \label{fig:teaser}
\end{figure}

Traditional \textit{stereo conversion} involves creating disparity maps from single images or sequences and then using them to generate the corresponding stereo pair for the other eye, creating the illusion of depth for stereoscopic viewing.
Recently, many deep learning-based methods~\citep{xie2016deep3d,wang2019web,Shih3DP20,watson2020learning,Ranftl2022} are primarily proposed for image-based stereo conversions, aiming to improve disparities and enhance inpainting effectiveness on occluded areas.
Unlike image data, video data provides additional temporal information, which can yield more detailed disparities and occlusion insights by leveraging information across frames. 
% To convert a film to stereo, dynamic objects and rapid scene changes pose significant challenges, requiring robust algorithms that can handle temporal consistency effectively.
% To handle video inputs, \cite{Chen2019} directly employs a UNet~\cite{ronneberger2015u}-shaped, 3D video-to-video translation model to generate right view from left view sequences.
To handle video inputs, \citet{Chen2019} synthesizes right-view video sequences by estimating a displacement map to move each pixel to a new location, with a 3D DenseNet.
Temporal3D~\citep{zhang2022temporal3d} compromises to use three adjacent left-view frames to predict the single right-view of the middle frame.
% Methods such as \textit{NVDS}~\cite{wang2023neural} are focused on producing temporal-consistent depth estimations from video inputs.
% To the best of our knowledge, a robust end-to-end stereo conversion framework for handling long video sequences does not exist.
Based on our analysis, current stereo conversion frameworks for video sequences are not robust and have several drawbacks. We believe the area is underexplored and there is a large room for improvement. At the same time, we believe the topic will gain in importance due to recent efforts to manufacture stereo displays, e.g., from Apple and Magic Leap.

% Methods such as \cite{Chen2019,zhang2022temporal3d,wang2023neural} propose different methods to estimate temporal-consistent disparity maps from a video sequence to compute the right view video.
% These approaches operate under the ill-posed assumption that occlusions are neglected, and all pixels in the left view are sufficient for generating the right view.
% To enhance the generation of coherent right-view videos, we aim to develop an end-to-end framework that effectively utilizes the temporal dimension.

% Obviously, in terms of video-based stereo conversion, these methods often lack the capability to produce coherent videos with consistent frame-to-frame transitions. 

% Furthermore, dynamic objects and rapid scene changes pose significant challenges, requiring robust algorithms that can handle temporal consistency effectively. Therefore, we aim to build an end-to-end framework to address the challenge of video stereo synthesis.

% Different from image-based methods, video input features motion along the temporal dimension that may further assist maintaining semantic consistency \wrt the input videos and representing occluded parts.

We introduce \textit{ImmersePro}, a novel approach designed specifically for video stereo conversion that utilizes the contextual information available across video frames to enhance stereo disparity consistency across the temporal dimension.
For doing so, we collectively build a large-scale stereo movie dataset, \textit{Youtube-SBS}, with over 7 million stereo pairs from a collection of stereo movies, game films, and music videos.
Due to the absence of ground truth disparities, we propose to use implicit disparities to guide the generation of \textit{layered disparities}, which outperforms the explicit disparity guidance (\eg a depth estimation model) that was commonly used in previous work. We propose to use a \textit{layered disparity} representation that refers to a stack of disparity maps corresponding to one image. Each pixel that appears in the image can be reused multiple times, avoiding creating black holes after the warping operation.
This approach ensures that the generated stereo parts strictly adhere to the semantics of the input video, minimizing the need for improvisation and thus preserving the original narrative and visual intent. 
As a result, \textit{ImmersePro} not only maintains the semantic integrity of the original video but also intelligently infers the geometry of occluded areas, enabling consistent right-view generation.
% Unlike existing methods that may perform inconsistently when applied to videos, \textit{ImmersePro} aims to leverage the inherent continuity in videos.
As shown in~\Cref{fig:teaser}, previous methods may generate artifacts such as texture misalignment or object deformation, whereas our \textit{ImmersePro} can keep the semantic integrity from the left-view image.
% Our method presents a significant advancement in stereo video generation, offering improved consistency and a more immersive viewing experience.
Our main contributions are as follows:
\begin{itemize}[leftmargin=*]
    \item We introduce the \textit{YouTube-SBS} dataset, an extensive collection of stereo videos sourced from YouTube, featuring over 7 million stereo pairs. This dataset fills the gap to serve as a benchmark for training and evaluating stereo video generation models.
    \item We introduce \textit{ImmersePro}, specifically tailored for converting single-view videos into stereo videos using \textit{layered disparity} warping via implicit disparity guidance. Compared to the best competitor stereo-from-mono we quantitatively improve the results by 11.76\% (L1), 6.39\% (SSIM), and 5.10\% (PSNR).
\end{itemize}

\section{Background}

We discuss previous stereo conversion methods and stereo datasets in this section.

\subsection{Stereo Conversion Methods}

\textbf{Image-Based Stereo Conversion.}
Deep3D~\citep{xie2016deep3d} relaxes the disparity map into a multi-layer probabilistic map and then multiplies it with several horizontally shifted copies of the input image, which relaxes the non-differentiable warping operation.
\citet{watson2020learning} used a warping-and-inpainting framework, which creates stereo training pairs from single RGB images to improve the modern monocular depth estimators.
However, a non-differentiable strategy is used and the inpainting randomly selects the texture from the training set.
Apart from using pretrained depth estimation models, \citet{wang2019web,Ranftl2022} use FlowNet2.0~\citep{IMKDB17} to estimate optical flows as ground truth disparities.
StereoDiffusion~\citep{wang2024stereodiffusion} proposes a training-free approach to generate stereo pairs by directly warping the latent space of diffusion models. It requires inversion methods to produce the latents to generate the stereo pair of a given image. The fine details of the resulting photo may vary due to the direct modification of the latent space.
\citet{Shih3DP20} proposed a layered depth inpainting method that generates a 3D representation by intelligently estimating and filling depth information, particularly in areas where it is missing or uncertain.
Our work does not rely on explicit disparity computation, with the additional consideration of the context within video frames.

\noindent\textbf{Video-Based Stereo Conversion.} 
% Video data, with its temporal dimension, provides additional temporal-coherent insights into missing areas that are not apparent in static images.
\citet{Chen2019} adopts a reconstruction-based approach by using a 3D DenseNet to estimate the disparity map of an input sequence.
\textit{Temporal3D}~\citep{zhang2022temporal3d} estimates the middle frame using three adjacent frames, with the output being a weighted sum of three disparity-warped images.
Additionally, methods such as \textit{NVDS}~\citep{wang2023neural} may be adopted for consistent depth estimations across video frames.
However, those methods assume the pixels within the left image are adequate for the right image.
\citet{mehl2024stereo} adopted the warping-inpainting approach with a pretrained depth estimation method (\ie MiDaS~\citep{birkl2023midas}) for warping and inpainting with multiple adjacent frames.
Still, this method relies on a single frame depth estimation model that can likely break the temporal consistency between frames. In this work, we propose an end-to-end video stereo conversion method based on implicit disparity guidance across the temporal dimension.

% Though a consistency check has been performed, the inaccurate disparity may still exist, resulting in accumulated errors.

% \begin{figure*}
%      \centering
%      \begin{subfigure}[b]{0.32\textwidth}
%          \centering
%          \includegraphics[width=.49\textwidth]{example-image-a}
%          \includegraphics[width=.49\textwidth]{example-image-a}
%          \includegraphics[width=.49\textwidth]{example-image-a}
%          \includegraphics[width=.49\textwidth]{example-image-a}
%          \caption{$<10\%$}
%      \end{subfigure}
%      \hfill
%      \begin{subfigure}[b]{0.32\textwidth}
%          \centering
%          \includegraphics[width=.49\textwidth]{example-image-b}
%          \includegraphics[width=.49\textwidth]{example-image-b}
%          \includegraphics[width=.49\textwidth]{example-image-b}
%          \includegraphics[width=.49\textwidth]{example-image-b}
%          \caption{$<20\%$}
%      \end{subfigure}
%      \hfill
%      \begin{subfigure}[b]{0.32\textwidth}
%          \centering
%          \includegraphics[width=.49\textwidth]{example-image-c}
%          \includegraphics[width=.49\textwidth]{example-image-c}
%          \includegraphics[width=.49\textwidth]{example-image-c}
%          \includegraphics[width=.49\textwidth]{example-image-c}
%          \caption{$<30\%$}
%      \end{subfigure}
%         \caption{Sample stereo images and their consistency checked results.}
%         \label{fig:three graphs}
% \end{figure*}

\subsection{Stereo Datasets}

There are limited resources on video-based stereo datasets.
Sintel~\citep{Butler:ECCV:2012} contains 1064 synthetic stereo images with accurate disparities.
KITTI~\citep{Menze2015CVPR} offers 8.4K frames captured from the real world for autonomous driving.
\citet{wang2019web} introduces a \textit{WSVD} dataset and proposes to use optical flow as disparities as ground truth for supervision.
Similarly, \citet{Ranftl2022} collected a private 3D movie dataset and extracted ground truth disparities by estimated optical flows to improve depth estimation.
Since different levels of stereoscopic effects may exist for different purposes of a dataset, a movie-specific benchmark dataset is preferable. \citet{Ranftl2022} is the only relevant dataset but it is built on top of real movies with intellectual property right issues. Therefore, we propose a benchmark stereo dataset that contains publicly available content.

\begin{table}[h]
    \scriptsize
    \centering
    \begin{tabular}{c|cccc}
        \toprule
         Dataset & content & GT depth & available & No. frames  \\
         \midrule
         KITTI~\citep{Menze2015CVPR}  & autonomous driving  & metric & Y & 8.4K \\
         % ReDWeb~\cite{xian2018monocular}  & & NA & Y & 3600 \\
         WSVD~\citep{wang2019web}  & mixed & NA & Y & 1.5M \\
         3D Movies~\citep{Ranftl2022} & movies & NA & N & 75K \\
         Sintel~\citep{Butler:ECCV:2012} & synthetic & metric & Y & 1064 \\
         \midrule
         Youtube-SBS & movies & NA & Y & 7M \\
         \bottomrule
    \end{tabular}
    \caption{Relevant datasets.}
    \label{tab:my_label}
\end{table}

\section{Youtube-SBS}

We aim to set up a large-scale publicly accessible benchmark dataset.
The direct collection of 3D movies often encounters legal challenges to publish as an open-source dataset. Therefore, we present \textit{Youtube-SBS}, an open-source dataset collected from YouTube. This dataset contains over 400 3D side-by-side (SBS) videos. With a particular interest in stereo movies, our dataset primarily consists of movie trailers, game films, and music videos. We explicitly excluded 360-degree virtual reality videos and gameplay videos (that contain user interfaces).
To ensure accessibility for future research, we select videos that (1) have existed for at least one year, and (2) from accounts that have at least 500 followers.
This curated selection includes 423 videos at a standard resolution of 1920x1080.
During the frame extraction, as some videos include a non-stereo intro section, we skip the first 600 frames to capture valid stereo pairs.

To measure the general stereo effects of our dataset, we propose to compute a metric that evaluates the left-right consistency of the disparity. 
For a stereo pair with subtle stereo effects, the disparity maps for the left and right images should be almost symmetrical with one another. That is, a point in the left image should have a corresponding point in the right image at the same row but shifted horizontally according to the disparity. For large stereo effects there is an increasing number of occluded and disoccluded areas. In these regions, the right image can no longer be reconstructed from the left image with simple warping (and vice versa).
% This check compares the disparities calculated from both the left and right views to identify inconsistencies that could indicate errors in the stereo data, such as occlusions.
% We experimented with several methods to estimate disparity.
% First, stereo-matching techniques such as STTR~\cite{li2021revisiting} are commonly trained on small datasets, so they suffer from limited generalizability to large-scale datasets.
% RAFT-Stereo~\cite{lipson2021raft} failed the consistency checks since the model is trained for left-to-right prediction, leading to significant discrepancies compared to right-to-left disparities. As we tested further, the optical flow using 
To compute our metric, we use the optical flow method RAFT~\citep{teed2020raft}. We also evaluated STTR~\citep{li2021revisiting} and RAFT-Stereo~\citep{lipson2021raft}, but these two methods produced worse results. Note that high consistency means that the left-to-right optical flow $F_{l\rightarrow r}$ and right-to-left optical flow $F_{r\rightarrow l}$ are the negative of each other.
We calculate the consistency $\varepsilon$ as follows: 
\begin{equation}
    \label{eq:consistency}
    \mathcal{E}_p = ||F_{l\rightarrow r}(p)) + F_{r\rightarrow l}(p+F_{l\rightarrow r}(p))||,
\end{equation}
where $p$ is the pixel position of a frame.
We provide a breakdown to demonstrate consistency metric in~\Cref{tab:dataset_consistency} to present the general stereo effects of the dataset.
We compute occluded areas with $\sum_{p}1(\mathcal{E}_{p} > \epsilon)$.
We use $\epsilon=4$ for improved stability on RAFT-computed optical flows.
We present a visual demonstration of different levels of stereo effects in~\Cref{fig:dataset}.
% where we compute the percentage of the errors in optical flows. At test time, the false prediction of optical flows may exist. Therefore, we do not exclude additional frames by a flow-based consistency check. 

\begin{table}[h]
    \centering
    \begin{tabular}{c|cccc}
        \toprule
        occluded area & $< 10\% $& $< 20\%$ & $< 30\%$ & $< 40\%$ \\
        \midrule
        Percentage & 71.27\% & 84.60\% & 91.30\% & 94.71\% \\
        \bottomrule
    \end{tabular}
    \caption{Flow-based consistency check results. Most frames present subtle stereo effects in the dataset.}
    \label{tab:dataset_consistency}
\end{table}

% In addition, since optical flows are used in our work, this experiment also assists in getting an idea about the robustness of the optical flow algorithm.

% However, we provide a breakdown to demonstrate our dataset.

\section{Method}

A stereo video sequence $I=\{I^l,I^r\}$ contains left and right video sequences of $I^l\in R^{T\times H\times W \times 3}$ and $I^r\in R^{T\times H\times W \times 3}$, respectively.
We use $T,H,W$ to denote the video sequence length, video height, and video width, respectively.
% With a sequence length of $T$, for each timestep $t\in \{0,\dots,T\}$, frames $I^l_t$ and $I^r_t$ are shaped as $H\times W \times 3$.
We aim to predict a right video sequence $\hat{I}^r$ based on the input left video sequence $I^l$ to make $\hat{I}=\{I^l,\hat{I}^r\}$ presents similar stereo effects as ${I}$.

\begin{figure*}
    \centering
    \includegraphics[width=\linewidth]{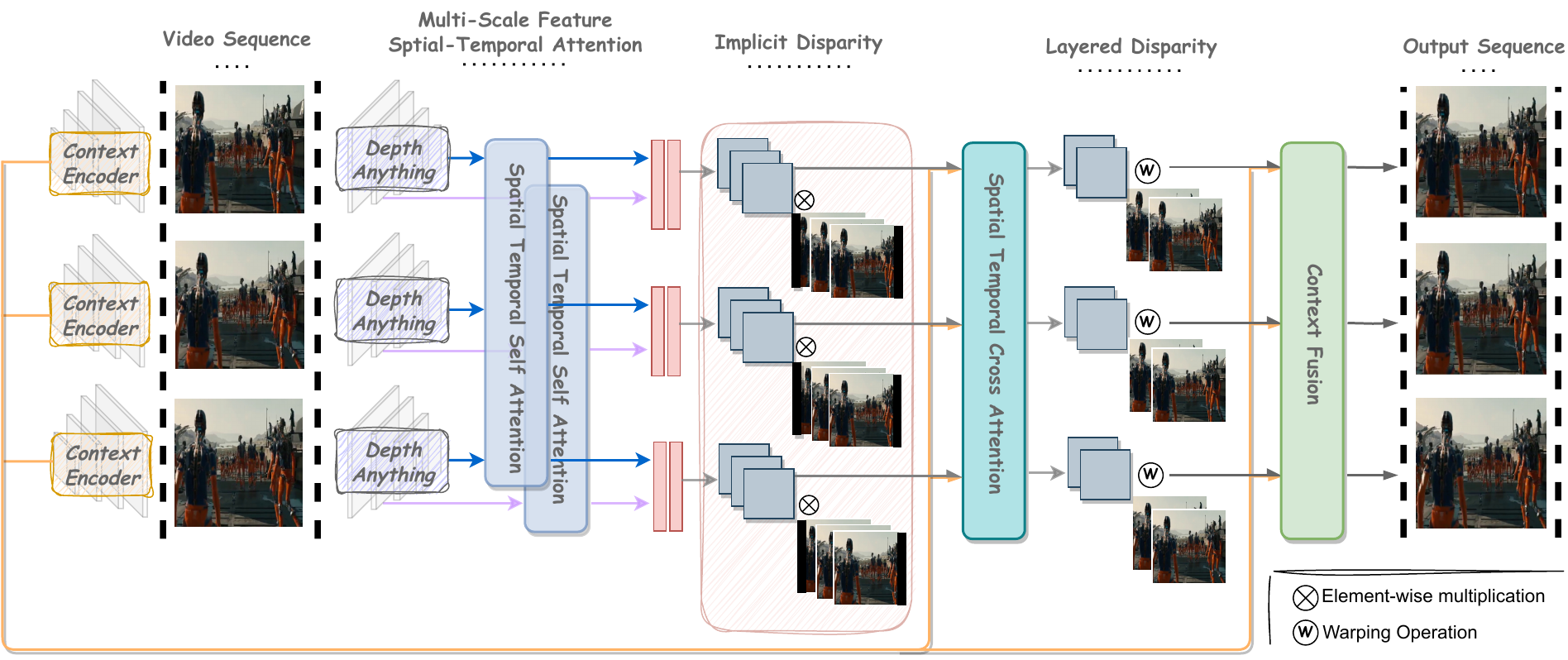}
    \caption{Illustration of ImmersePro framework. Our network contains six parts: (1) dual-branch feature extractors for extracting disparity features and context features (\cref{sec:dual_branch}), (2) multi-scale spatial-temporal self-attention to refine disparity features (\cref{sec:attention}), (3) implicit disparity to generate stereo images without explicit disparities (\cref{sec:implicit_disp}), (4) spatial-temporal cross attention block to inject contextual information into the implicit disparity features (\cref{sec:attention}), (5) layered disparity to obtain the estimated right view video sequences (\cref{sec:layered_disp}), and (6) context fusion to enrich the estimated right view video sequences with detailed semantic information (\cref{sec:context_fusion}).}
    % \Description{Illustration of ImmersePro framework}
    \label{fig:main}
\end{figure*}

% To better capture the motion information across frames, we extract the forward and backward optical flows, denoted as $F^f=\{F^f_t = F_{t\rightarrow t+1}\in \mathbb{R}^{H\times \frac{W}{2}\times 2}\}^{T-1}_{t=1}$ and $F^b=\{F^b_t = F_{t+1\rightarrow t}\in \mathbb{R}^{H\times \frac{W}{2}\times 2}\}^{T-1}_{t=1}$ from the left video sequences $I^l$.
% In general, we input left video sequence $I^l$ and its optical flows $F^f$ and $F^b$ to predict the corresponding right video sequence $\hat{I}=\{\hat{I}_t^r\in \mathbb{R}^{H\times \frac{W}{2}\times 2}\}^T_{t=1}$.
% We aim to predict the corresponding right video sequence $\hat{I}^r=\{\hat{I}_t^r\in \mathbb{R}^{H\times \frac{W}{2}\times 2}\}^T_{t=1}$ based on input left video sequence $I^l$.
As shown in~\Cref{fig:main}, our method compromises six stages. First, we use a dual branch architecture (\cref{sec:dual_branch}) that consists of a disparity branch and a context branch, to extract disparity and semantic features, respectively. Second, we apply spatial-temporal self-attention (\cref{sec:attention}) on each scale feature to achieve multi-frame awareness. Third, we fuse the multi-scale features to obtain implicit disparity features (\cref{sec:implicit_disp}). Fourth, we then use a spatial-temporal cross-attention module (\cref{sec:attention}) to inject contextual information into the implicit disparity features to obtain layered disparity features (\cref{sec:layered_disp}). Fifth, right-view video sequences can be estimated by warping through layered disparities. Finally, we enrich the estimated right-view sequences with a context fusion module.

% Our feature branch features a flow-guided bidirectional propagation (\Cref{sec:flow}) and a stereo feature conversion (\Cref{sec:stereo_conv}) module for offering the stereoscopic features. 
% Finally, a decoder that fuses features from both branches is used to reconstruct them to the final right video sequence.
% Aside from being self-coherent, the generated right video sequences need to strictly follow the semantics of the left video sequence. In addition, a meaningful improvisation of the potential occlusion is desired. 

\subsection{Dual Branch Architecture}
\label{sec:dual_branch}

We use a dual-branch architecture to enhance stereo video conversion by separately processing disparity and contextual information, as shown in~\Cref{fig:main}.
We employ a pretrained DepthAnything~\citep{yang2024depth} model for the disparity branch to extract disparity-oriented feature maps, while a context feature extractor with the same architecture from~\citet{zhou2023propainter,liCvpr22vInpainting}'s encoder is used to extract contextual semantic features.

The disparity branch operates on multiple scales, extracting features at $1/2$ and $1/4$ resolutions of the original input to capture detailed disparity information.
The disparity features are from the decoder of the model\footnote{We use the output from the neck of the model, as implemented by  \url{https://github.com/huggingface/transformers}.}.
% flow-guided feature propagation to maintain temporal consistency across frames, crucial in dynamic scenes with significant motion. Additionally, 
This branch utilizes spatial-temporal self-attention modules (\cref{sec:attention}) to prioritize relevant spatial and temporal details on different scales, ensuring that the model focuses on areas with significant disparity changes or movement.
After combining the multi-scale features into $1/2$ resolution with a fusion block, we apply softmax to these features to create a probability distribution that represents the implicit disparities.
The implicit disparity is used to select the appropriate pixels from a stack of the multiple horizontally shifted copies of the input image (\cref{sec:implicit_disp}). By encouraging accurate selection, these features implicitly represent the disparity for stereo conversion.

\begin{figure*}[t]
     \centering
     \begin{tblr}{colspec={Q[c]Q[c]Q[c]Q[c]Q[c]Q[c]Q[c]}, colsep = {.1pt}, rowsep = {.3pt}}
        \adjustbox{valign=m}{\scriptsize 
        \rot{\Tr{Input image}}}
        % {0 40.6cm 0 27.2cm}
            & \includegraphics[width=.18\textwidth,height=.13\textwidth,trim={0 67.8cm 0 0},clip, valign=c]{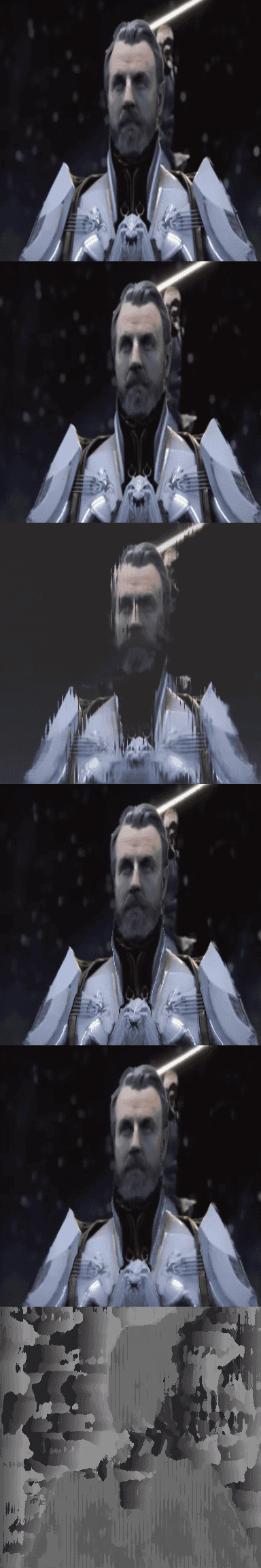}
            & \includegraphics[width=.18\textwidth,height=.13\textwidth,trim={0 67.8cm 0 0},clip, valign=c]{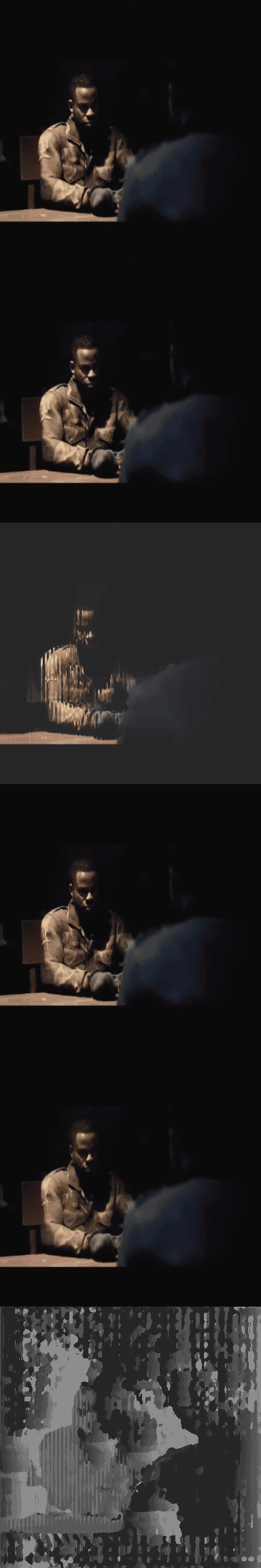}
            & \includegraphics[width=.18\textwidth,height=.13\textwidth,trim={0 67.8cm 0 0},clip, valign=c]{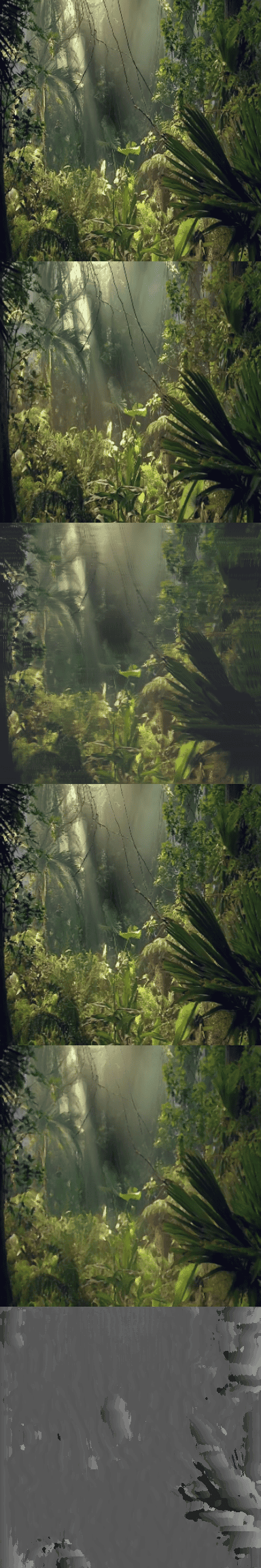}
            & \includegraphics[width=.18\textwidth,height=.13\textwidth,trim={0 67.8cm 0 0},clip, valign=c]{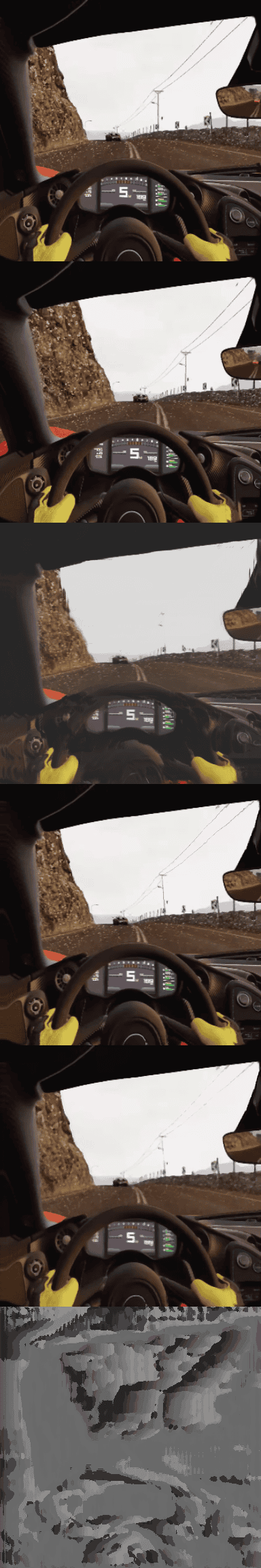}
            & \includegraphics[width=.18\textwidth,height=.13\textwidth,trim={0 67.8cm 0 0},clip, valign=c]{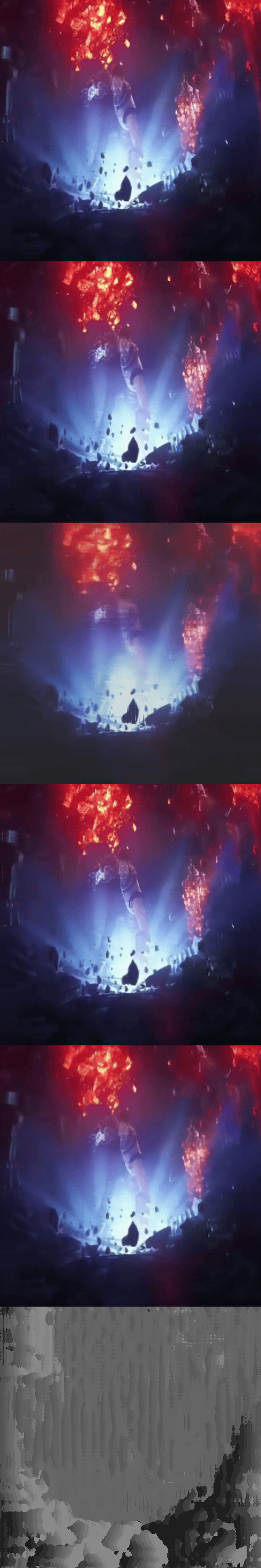} \\
        \adjustbox{valign=m}{\scriptsize \rot{\Tr{\makecell{implicit disparity}}}}
        % {0 40.6cm 0 27.2cm}
            & \includegraphics[width=.18\textwidth,height=.13\textwidth,trim={0 0 0 67.8cm},clip, valign=c]{misc/disparity/new_mask_37200.png}
            & \includegraphics[width=.18\textwidth,height=.13\textwidth,trim={0 0 0 67.8cm},clip, valign=c]{misc/disparity/new_mask_41000.png}
            & \includegraphics[width=.18\textwidth,height=.13\textwidth,trim={0 0 0 67.8cm},clip, valign=c]{misc/disparity/new_mask_46200.png}
            & \includegraphics[width=.18\textwidth,height=.13\textwidth,trim={0 0 0 67.8cm},clip, valign=c]{misc/disparity/new_mask_49600.png}
            & \includegraphics[width=.18\textwidth,height=.13\textwidth,trim={0 0 0 67.8cm},clip, valign=c]{misc/disparity/new_mask_49800.png} \\
        \adjustbox{valign=m}{\scriptsize \rot{\Tr{\makecell{\textbf{Output} w/\\implicit disparity}}}}
        % {0 40.6cm 0 27.2cm}
            & \includegraphics[width=.18\textwidth,height=.13\textwidth,trim={0 40.6cm 0 27.2cm},clip, valign=c]{misc/disparity/new_mask_37200.png}
            & \includegraphics[width=.18\textwidth,height=.13\textwidth,trim={0 40.6cm 0 27.2cm},clip, valign=c]{misc/disparity/new_mask_41000.png}
            & \includegraphics[width=.18\textwidth,height=.13\textwidth,trim={0 40.6cm 0 27.2cm},clip, valign=c]{misc/disparity/new_mask_46200.png}
            & \includegraphics[width=.18\textwidth,height=.13\textwidth,trim={0 40.6cm 0 27.2cm},clip, valign=c]{misc/disparity/new_mask_49600.png} 
            & \includegraphics[width=.18\textwidth,height=.13\textwidth,trim={0 40.6cm 0 27.2cm},clip, valign=c]{misc/disparity/new_mask_49800.png} \\
        \adjustbox{valign=m}{\scriptsize \rot{\Tr{\makecell{\textbf{Output} w/\\layered disparity}}}}
            & \includegraphics[width=.18\textwidth,height=.13\textwidth,trim={0 13.6cm 0 54.2cm},clip, valign=c]{misc/disparity/new_mask_37200.png}
            & \includegraphics[width=.18\textwidth,height=.13\textwidth,trim={0 13.6cm 0 54.2cm},clip, valign=c]{misc/disparity/new_mask_41000.png}
            & \includegraphics[width=.18\textwidth,height=.13\textwidth,trim={0 13.6cm 0 54.2cm},clip, valign=c]{misc/disparity/new_mask_46200.png}
            & \includegraphics[width=.18\textwidth,height=.13\textwidth,trim={0 13.6cm 0 54.2cm},clip, valign=c]{misc/disparity/new_mask_49600.png}
            & \includegraphics[width=.18\textwidth,height=.13\textwidth,trim={0 13.6cm 0 54.2cm},clip, valign=c]{misc/disparity/new_mask_49800.png} \\
        \adjustbox{valign=m}{\scriptsize \rot{\Tr{\makecell{GT right}}}}
            & \includegraphics[width=.18\textwidth,height=.13\textwidth,trim={0 54.2cm 0 13.6cm},clip, valign=c]{misc/disparity/new_mask_37200.png}
            & \includegraphics[width=.18\textwidth,height=.13\textwidth,trim={0 54.2cm 0 13.6cm},clip, valign=c]{misc/disparity/new_mask_41000.png}
            & \includegraphics[width=.18\textwidth,height=.13\textwidth,trim={0 54.2cm 0 13.6cm},clip, valign=c]{misc/disparity/new_mask_46200.png}
            & \includegraphics[width=.18\textwidth,height=.13\textwidth,trim={0 54.2cm 0 13.6cm},clip, valign=c]{misc/disparity/new_mask_49600.png}
            & \includegraphics[width=.18\textwidth,height=.13\textwidth,trim={0 54.2cm 0 13.6cm},clip, valign=c]{misc/disparity/new_mask_49800.png} \\
        % \adjustbox{valign=m}{\scriptsize \rot{\Tr{\makecell{GT right}}}}
        %     & \includegraphics[width=.16\textwidth,height=.13\textwidth,trim={0 54.2cm 0 13.6cm},clip, valign=c]{misc/wrong_stereo_step73200.png}
        %     & \includegraphics[width=.16\textwidth,height=.13\textwidth,trim={0 54.2cm 0 13.6cm},clip, valign=c]{misc/good_stereo_step19200.png}
        %     & \includegraphics[width=.16\textwidth,height=.13\textwidth,trim={0 54.2cm 0 13.6cm},clip, valign=c]{misc/good_stereo_step20000.png}
        %     & \includegraphics[width=.16\textwidth,height=.13\textwidth,trim={0 40.65cm 0 10.15cm},clip, valign=c]{misc/good_stereo_step21400_w_square.png}
        %     & \includegraphics[width=.16\textwidth,height=.13\textwidth,trim={0 54.2cm 0 13.6cm},clip, valign=c]{misc/good_stereo_step79000.png}
        %     & \includegraphics[width=.16\textwidth,height=.13\textwidth,trim={0 54.2cm 0 13.6cm},clip, valign=c]{misc/good_stereo_step83600.png}\\
    \end{tblr}
    \caption{Visual demonstration of the implicit disparity guidance. We can observe that \textit{(1)} the implicit disparity module tries to resolve the disparity from the given image, and \textit{(2)} our method can significantly rectify the error introduced by the implicit disparity estimations. Our method offers a significant improvement regarding clarity with less irregular texture deformation on the image.  The implicit disparity map contains multiple channels and we apply $argmax$ to obtain the visual output.}
    \label{fig:disp_diff}
\end{figure*}

Concurrently, we use a stack of convolution layers as the context encoder. We experimented with multiple encoder architectures and settled on the architecture without aggressive downsampling. The details for the context encoder are presented in~\Cref{sec:context_encoder}.
The context branch focuses solely on capturing texture information. This branch processes texture at $1/2$ the original resolution, aligning with the disparity branch's output.
Finally, with spatial-temporal cross-attention modules to fuse the implicit disparity and texture information, we apply a layered disparity warping (\cref{sec:layered_disp}) to obtain the final predicted right-view. 
% Next, 

% This creates disparity-guided right view whilst keeping the overall semantics of the left image.

% Finally, the features from the two branches are fused with additional spatial-temporal attention blocks to fine-tune the alignment and integration of disparity and contextual data before decoding to the final right-view videos.

\subsection{Spatial-Temporal Attention}
\label{sec:attention}
Video transformers have demonstrated excellent performances in video-based tasks such as video segmentation~\citep{duke2021sstvos}, video-text feature mapping~\citep{li2023svitt}, and video inpainting~\citep{liCvpr22vInpainting,zhou2023propainter}.
This work builds sparse video transformers on top of the ProPainter's version, considering the highly redundant and repetitive textures in adjacent frames.
We remove the mask guidance in the original model and use a temporal stride of 2 to avoid redundant key/value tokens within each transformer block and to improve the computational efficiency.
Aside from spatial-temporal self-attention, we also use spatial-temporal cross-attention to fuse features from different sources.

Given a video feature sequence $E_s \in \mathbb{R}^{T_s\times H_s \times W_s \times C}$, we first perform soft split~\citep{liu2021fuseformer} to generate patch embeddings $Z\in \mathbb{R}^{T_s\times M \times N \times C_z}$.
Subsequently, $Z$ is partitioned into $m \times n$ non-overlapping windows, yielding the partitioned embedding features $Z_w \in \mathbb{R}^{T_s\times m \times n \times h \times w \times C_z}$, where $m\times n$ denotes the number of windows and $h\times w$ denotes their size.
For self-attention transformer blocks, we obtain the query $Q$, key $K$, and value $V$ from $Z_w$ through three linear layers, respectively.
For cross-attention transformer blocks, we repeat the above process to obtain embeddings $Z_c \in \mathbb{R}^{T_s\times m \times n \times h \times w \times C_z}$ from another feature sequence $E_c \in \mathbb{R}^{T_s\times H_s \times W_s \times C}$. Note that $Z_c$ shares the same shape with $Z_w$. Then $Q$ is extracted from $Z_w$ whilst $K$ and $V$ are extracted from $Z_c$.
% Strategies such window expand~\cite{liu2021fuseformer} and global tokens~\cite{zhang2022flow} are applied.
For both self-attention and cross-attention mechanisms, the final embedding features are gathered using soft composition~\cite{liu2021fuseformer} for further processing.

\subsection{Implicit Disparity}
\label{sec:implicit_disp}

For stereo vision, different from common generative tasks, the generated right view requires a precise match to the input view with as little improvisation as possible.
The stereo pair of an image is commonly constructed by obtaining the disparity map to find the shifting distances of each pixel within the input view.
Assuming $d_{i,j}$ is the disparity value at pixel location $(i,j)$ in the left image, the corresponding pixel in the right image is:
\begin{equation}
    I^r_{i,j} = I^l_{i,j + d_{i,j}}
    \label{eq:disp}.
\end{equation}
It is typically a non-differentiable operation due to its piecewise nature. \citet{jaderberg2015spatial} propose to use sub-gradients for backpropagation through spatial transformations to handle such non-smooth operations, enabling differentiable warping.

% Xie \etal~
\citet{xie2016deep3d} proposed another approach to use a depth selection layer to align the output right view to the source input view structure. Subsequent works such as~\citet{zhang2019structure} follow a similar idea. We employ it as auxiliary supervision. We found this method to be suitable for guidance only. Directly using it to compute the output leads to blurry results.
\emph{Implicit disparity} predicts a probability distribution across possible disparity values $d$ at each pixel location. $p^d_{i,j}$, with $\sum_d p^d_{i,j}=1$, denotes the probability of pixel $(i,j)$ having disparity $d$.
We denote an image that is shifted by $d$ pixels horizontally as $I^d_{i,j}=I_{i,j-d}$. We then obtain the right-view pixel values as:
\begin{equation}
    \hat{I}^{aux}_{i,j} \sum_d = I^d_{i,j} p^d_{i,j}.
\end{equation}
where $\hat{I}^{aux}$ is the auxiliary predicted right view. We use $V^d_{i,j} = I^d_{i,j} D^d_{i,j}$ for subsequent computations. This approach estimates the stereo pair of a given image without an explicit disparity map, serving as a relaxation of the warping operation in \Cref{eq:disp}.
Without implicit disparity, our model can hardly converge as shown in~\Cref{sec:results}.
% This curated stereo conversion method regularizes the disparity range by tweaking the minimum and maximum values of $d$, mimicking actual stereo conversion behavior.
% In particular, this method approximates pixel-wise displacement by selecting the appropriate disparity levels for each pixel from a set of candidate depths.

\begin{figure*}[t]
     \centering
     \begin{subfigure}[b]{.26\textwidth}
         \centering
         \includegraphics[width=\textwidth,height=.8\textwidth]{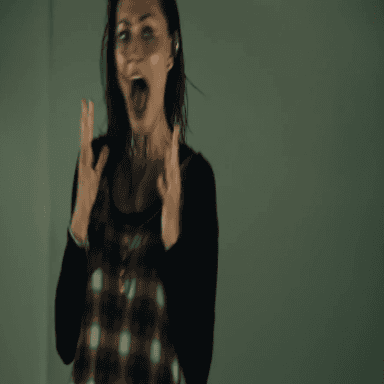}
         \caption{Left image}
     \end{subfigure}%
     \begin{subfigure}[b]{.47\textwidth}
        \centering
        \begin{subfigure}[b]{\textwidth}
            \centering
            \includegraphics[width=.7\textwidth]{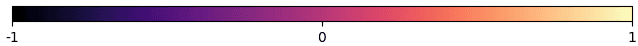}
        \end{subfigure}
        \begin{subfigure}[b]{\textwidth}
             \begin{subfigure}[b]{.49\textwidth}
                 \centering
                 \includegraphics[width=\textwidth,height=.8\textwidth]{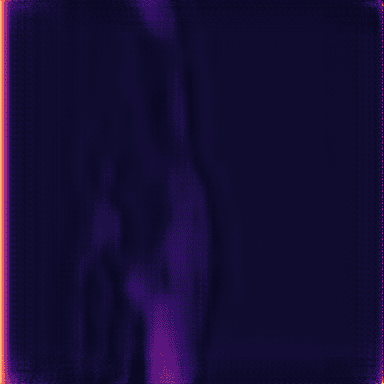}
                 \caption{$1^{st}$ layer}
                 \label{fig:layered-4}
             \end{subfigure}%
             \begin{subfigure}[b]{.49\textwidth}
                 \centering
                 \includegraphics[width=\textwidth,height=.8\textwidth]{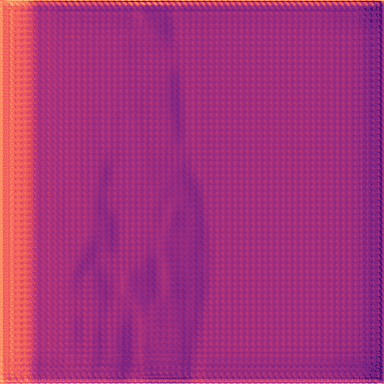}
                 \caption{$3^{rd}$ layer}
                 \label{fig:layered-5}
             \end{subfigure}
        \end{subfigure}
     \end{subfigure}%
     \begin{subfigure}[b]{.26\textwidth}
         \centering
         \includegraphics[width=\textwidth,height=.8\textwidth]{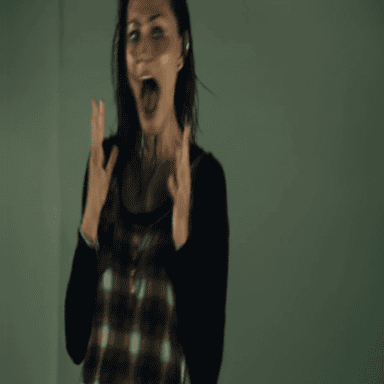}
         \caption{Predicted right}
     \end{subfigure}

      \caption{Visual demonstration of our layered disparity representation. We show the $1^{st}$ and $3^{rd}$ disparity layers in \Cref{fig:layered-4,fig:layered-5}. We denote darker colors as moving to the right and lighter colors as moving to the left. We use 7 layers in total while we found the $1^{st}$ and $3^{rd}$ layers contribute to the right-view generation most. We observe that the $1^{st}$ layer aims to warp the majority of the pixels to the right to their correct right-view location while the $3^{rd}$ layer moves pixels to the left to fill the resulting holes, e.g. near the left border.}
        \label{fig:layered_disp}
\end{figure*}

% To take advantage of these regularized and implicitly learned disparity features, we propose to use th
% Otherwise, this operation may distort the resulting image, leading to potential numerical issues.
% Thus, the disparity feature $D^d_{i,j}$ 

% $\sum_d V^d_{i,j}$ as an intermediate output to guide the network to assist in learning the disparities for stereo video synthesis

% In this work, we take the summation $\sum_d V^d_{i,j}$ as an intermediate output to guide the network to assist in learning the disparities for stereo video synthesis.

% We therefore supplement with additional context information to mitigate the information loss.

\begin{algorithm}[h]
\SetAlgoLined
    \PyComment{number\textunderscore layered\textunderscore disparity: the number of disparity layers.} \\
    \PyComment{warped\textunderscore output: `BDTCHW`. A stack of images warped by layered disparities. D is the number of disparity layers.} \\
    \PyComment{warped\textunderscore mask: `BDTCHW`. A stack of masks warped by layered disparities. D is the number of disparity layers.} \\
    \PyCode{layered\textunderscore mask = zeros\textunderscore like(output\textunderscore mask)} \\
    \PyCode{total\textunderscore mask = zeros\textunderscore like(output\textunderscore mask)} \\
    \PyCode{for i in range(number\textunderscore layered\textunderscore disparity):} \\
    \Indp   % start indent
        \PyCode{if i == 0:} \\
        \Indp   % start indent
            \PyCode{layered\textunderscore mask[:, i] = warped\textunderscore mask[:, i]} \\
            \PyCode{total\textunderscore mask[:, i] = warped\textunderscore mask[:, i]} \\
        \Indm % end indent, must end with this, else all the below text will be indented
        \PyCode{else:} \\
        \Indp   % start indent
            \PyCode{total\textunderscore mask[:, i] = logical\textunderscore or(warped\textunderscore mask[:, i], layered\textunderscore mask[:, i - 1])} \\
            % \PyCode{layered\textunderscore mask[:, i] = total\textunderscore mask[:, i] - warped\textunderscore mask[:, i - 1]} \\
            \PyCode{layered\textunderscore mask[:, i] = torch.logical\textunderscore and((1 - total\textunderscore mask[:, i]), warped\textunderscore mask[:, i - 1])} \\
            % \PyCode{layered\textunderscore mask[:, i] = torch.logical\textunderscore and((1 - total\textunderscore mask[:, i]), output\textunderscore mask[:, i - 1])} \\
        \Indm % end indent, must end with this, else all the below text will be indented
    \Indm % end indent, must end with this, else all the below text will be indented
    \PyCode{output = layered\textunderscore mask * warped\textunderscore output} \\
\caption{Synthesis from layered disparities.}
\label{algo:layered_disp}
\end{algorithm}

\subsection{Layered Disparity}
\label{sec:layered_disp}
% Previous works such as 3D photograph~\cite{Shih3DP20} use depth map layered depth image~\cite{shade1998layered}

The \textit{implicit disparity} is a summation-based approach that computes pixel colors as a blend of other pixel colors, weighted by the estimated probabilities. This may produce good results with a correct estimation, but it may introduce artifacts such as blurring if the estimation is inaccurate. The final output visually improves if each pixel location is selected from a set of candidate disparity layers, rather than blending all the layers.
The proposed \emph{Layered Disparity} uses a smaller stack of candidate layers, and each layer represents disparity information. 
% To avoid unexpected image distortion by the varying probabilities, we propose to take advantage of these regularized disparity features as implicit guidance to obtain layered disparity.
Therefore, our layered disparity representation is a stack of disparity maps. We use a differentiable warping~\citep{jaderberg2015spatial} operation to warp the input image to an output image. While a single disparity map already defines a solution to the problem, there may be problems due to occlusion and disocclusion artifacts. These problems can then be fixed by other layers.
% , to avoid the issue mentioned above of inaccurate pixel value predictions from the varying probabilities.
Our approach avoids the mentioned blending problem. Meanwhile, we maximize the reuse of pixel information within the image while at the same time avoiding generating image holes.

We use implicit disparity $V^d_{i,j}$ as a guidance to generate layered disparities.
First, we employ three \textit{Conv-ReLU} blocks to refine the $V^d_{i,j}$ to shrink them from $\mathbb{R}^{T_s \times H\times W \times D}$ into $\mathbb{R}^{T_s \times \frac{H}{2}\times\frac{W}{2} \times D}$, where $D$ is the number of stacked disparities. We then apply the spatial-temporal cross-attention process, as mentioned in~\Cref{sec:attention}. With the attention-applied features, a deconvolution operation and three \textit{Conv-ReLU} blocks are used to obtain the final layered disparity $LD_{i,j}^d$. Here, $d=7$ since we use 7 disparity layers in our work. We then apply the differentiable warping operation with the layered disparity to obtain layered warped images $\hat{I}_{i,j}$ and masks $\hat{M}_{i,j}$, respectively. We select pixel values according to the layered masks as in~\Cref{algo:layered_disp}. As shown in~\Cref{fig:disp_diff}, our proposed approach significantly improved the visual quality compared to the output from the implicit disparity layer. \Cref{fig:layered_disp} visualizes an example of learned disparity maps from the proposed layered disparity representation.

\subsection{Context Fusion}
\label{sec:context_fusion}

The final stage of our network focuses on enriching semantic details while maintaining the learned right-view structure. 
The context fusion module integrates semantic and disparity features from a video sequence by concatenating the encoder feature map with layered disparity features to form a fused representation. These fused features are then processed through spatial-temporal attention (\cref{sec:attention}), enabling global context awareness.
We apply spatial-temporal attention modules at 1/2 the original resolution, as mentioned in~\cref{sec:dual_branch}.
To retain structural integrity, a residual connection reintroduces the refined transformer output into the original fused feature map.
We then apply a deconvolution to obtain a texture map in the original resolution, then enrich the texture map by three \textit{Conv-ReLU} blocks. Next, the module supplements the layered disparity-warped images from~\cref{sec:layered_disp} with the enriched feature map. To be specific, a median blur with $3\times 3$ kernels is first applied to the warped images to reduce noise and improve local smoothness before concatenating them with the enriched feature map. A semantic residual is then derived by passing this combined map through three \textit{Conv-ReLU} blocks. The final output is produced by combining the blurred image with the semantic residual.
This approach ensures that the final result maintains sharp textures while preserving structural consistency, achieving a balance between local detail and global coherence.

\section{Results}

We implement our method using Pytorch and train on four NVIDIA A100 (80G) GPUs for 50,000 iterations (approx. 3 days). Models are trained for 40,000 iterations for our ablations. At training time, we first resize the input sequence to $422\times 422$ and then randomly crop the resized video sequence to $384\times 384$. Each input sequence contains 8 frames.
We use L1 loss during training to encourage an accurate reconstruction of the right-view images using both implicit and layered disparities. In addition, an LPIPS~\citep{zhang2018perceptual} loss is used for better reconstruction results.
An AdamW~\citep{loshchilov2017decoupled} optimizer is used. We use $3e-5$ learning rate while image losses are computed within the range of $(-127.5, 127.5)$.
% We provide training details in~\Cref{sec:training}.
We evaluated our method on our test set which includes 43 video sequences with 558K frames.

\begin{table}[t]
    \centering
    \small
    \begin{tabular}{l|c c c c}
        \toprule
        & L1 $\downarrow$ & SSIM $\uparrow$ & PSNR $\uparrow$\\
        \midrule
         Deep3D & 0.2215 & 0.1935 & 11.9089\\
         3D Photo & 0.1069 & 0.3463 & 16.3658 \\
         Stereo Diffusion & 0.0816 & 0.4651 & 18.6684 \\
         stereo-from-mono & 0.0646 & 0.5685 & 20.7788  \\
        \midrule
        Ours w/o implicit disparity $*$ & n/a & n/a & n/a \\
        Ours w/o layered disparity & 0.0885 & 0.4717 & 19.0523 \\
        Ours w/o attention blocks &  0.0593 & 0.5894 & 21.4162 \\
         Ours w/o context fusion & \cellcolor{yellow!25} 0.0588 &  \cellcolor{yellow!25} 0.5959 & \cellcolor{yellow!25} 21.6649 \\
        \midrule
        % Ours (MiDaS) & 0.0590 & 0.6014 & 21.6572 \\
        Ours & \cellcolor{green!25} 0.0570 & \cellcolor{green!25} 0.6048 & \cellcolor{green!25} 21.8387 \\
        \bottomrule
    \end{tabular}
    \caption{Benchmark results. The best and second-best results are highlighted in green and yellow, respectively. $*$ indicates the model is not converged.}
    \label{tab:main_table}
\end{table}

% Our final model achieves the highest SSIM score of 0.6030, demonstrating superior structural preservation of the right-view images. In comparison, the best-performing baseline, stereo-from-mono, attains an SSIM of 0.5685. Other baseline methods, such as Deep3D and 3D Photo, perform significantly worse with SSIM scores of 0.1935 and 0.3463, respectively.

% The importance of each component in our model is illustrated through ablation experiments. Removing attention blocks leads to a slight drop in SSIM to 0.5894, while omitting the layered disparity further reduces the SSIM to 0.4717. Removing temporal normalization results in a competitive SSIM of 0.5959 but underperforms compared to the full model. Notably, removing implicit disparity caused the model to fail to converge, suggesting the critical role of this component in our method.

\subsection{Comparison with State-of-art models}
\label{sec:results}
\textit{Benchmark methods.} We compare our method with three state-of-the-art methods including Stereo-from-mono~\citep{watson2020learning}, 3D Photography~\citep{Shih3DP20}, and StereoDiffusion~\citep{wang2024stereodiffusion}. Note that those methods are designed for image-based stereo conversion purposes. We are not aware of any open-source implementations for video stereo conversion. We use official implementations for the selected methods.

\noindent\textit{Benchmark settings.}
Due to the high runtime of those methods (especially for StereoDiffusion which is required to perform inversion~\citep{mokady2023null} for each image), we compare those methods with a subsampled dataset every 3 seconds (72 frames). At test time, we process 8 frames as input where the last 2 frames are taken as reference frames. We use widely employed L1, SSIM, and PSNR to evaluate the quality of the generated stereo pairs.

\begin{figure}[t]
     \centering
     \begin{tblr}{colspec={Q[c]Q[c]Q[c]Q[c]Q[c]}, colsep = {0pt}, rowsep = {0pt}}
        % \adjustbox{valign=m}{\rot{\Tr{\makecell{GT left}}}}
        \scriptsize Left Image & \scriptsize \makecell{Predicted \\ L2R Disparity} & \scriptsize \makecell{Predicted \\ Red-Blue Stereo} & \scriptsize \makecell{GT \\ L2R Disparity} & \scriptsize \makecell{GT \\ Red-Blue Stereo} \\
            
        \includegraphics[width=.20\textwidth,height=.16\textwidth,clip,valign=c]{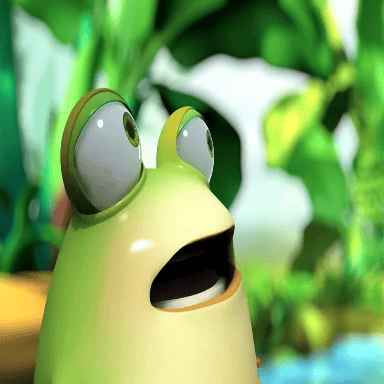}
        &
        \includegraphics[width=.20\textwidth,height=.16\textwidth,clip,valign=c]{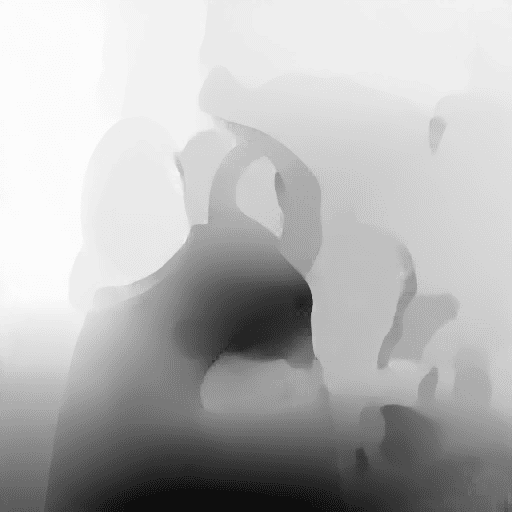}
        &
        \includegraphics[width=.20\textwidth,height=.16\textwidth,clip,valign=c]{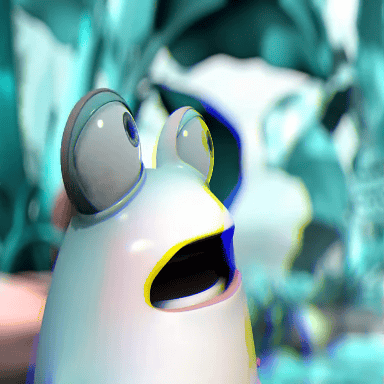} 
        &
        \includegraphics[width=.20\textwidth,height=.16\textwidth,clip,valign=c]{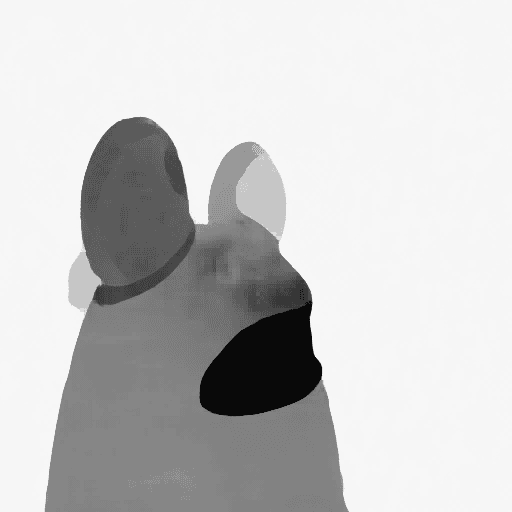} 
        &
        \includegraphics[width=.20\textwidth,height=.16\textwidth,clip,valign=c]{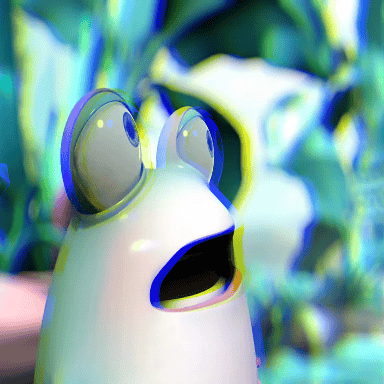} \\
        \includegraphics[width=.20\textwidth,height=.16\textwidth,clip,valign=c]{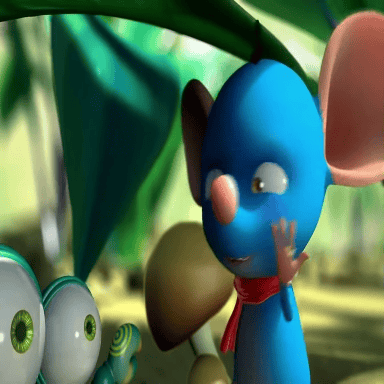}
        &
        \includegraphics[width=.20\textwidth,height=.16\textwidth,clip,valign=c]{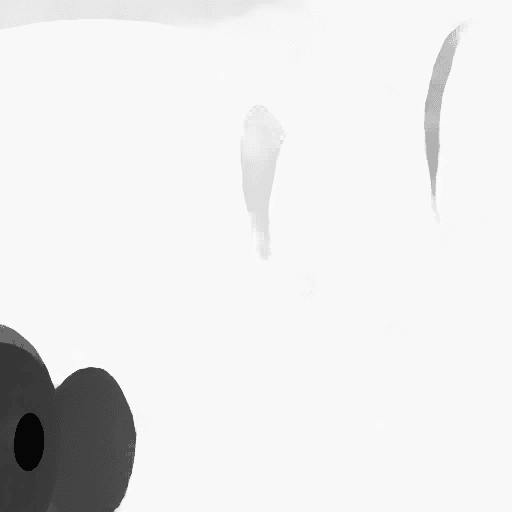}
        &
        \includegraphics[width=.20\textwidth,height=.16\textwidth,clip,valign=c]{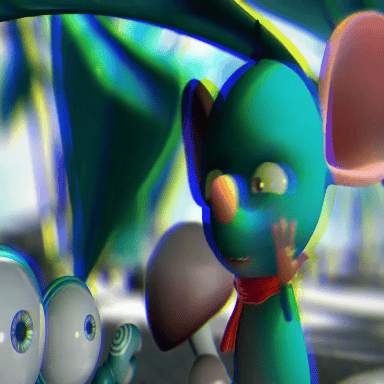} 
        &
        \includegraphics[width=.20\textwidth,height=.16\textwidth,clip,valign=c]{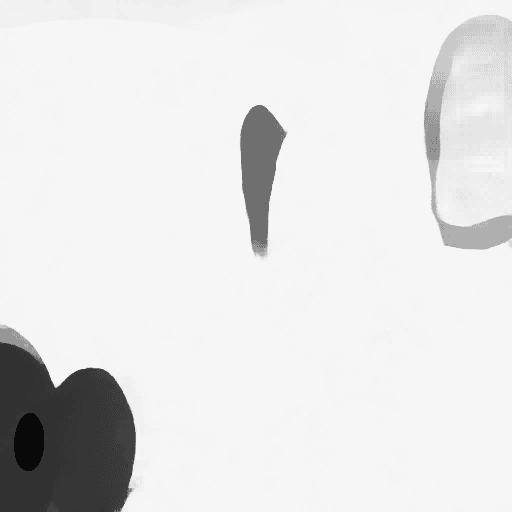} 
        &
        \includegraphics[width=.20\textwidth,height=.16\textwidth,clip,valign=c]{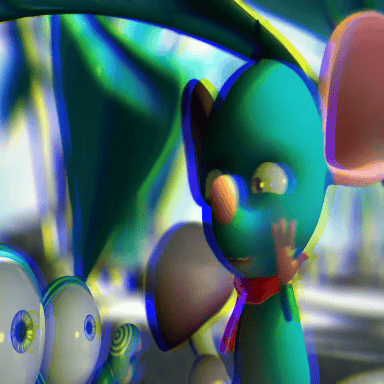} \\
            
        \includegraphics[width=.20\textwidth,height=.16\textwidth,clip,valign=c]{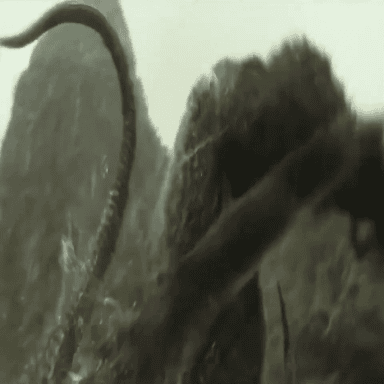}
        &
        \includegraphics[width=.20\textwidth,height=.16\textwidth,clip,valign=c]{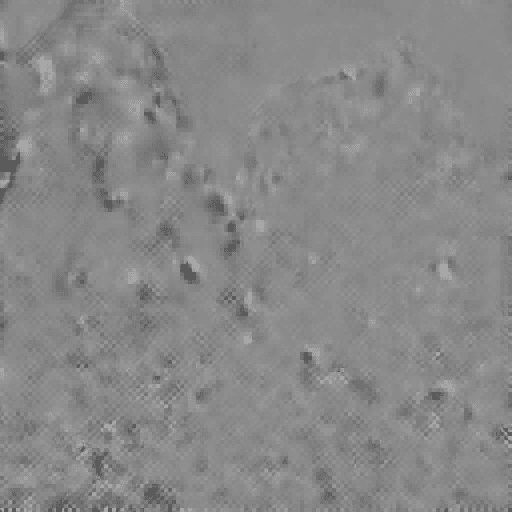}
        &
        \includegraphics[width=.20\textwidth,height=.16\textwidth,clip,valign=c]{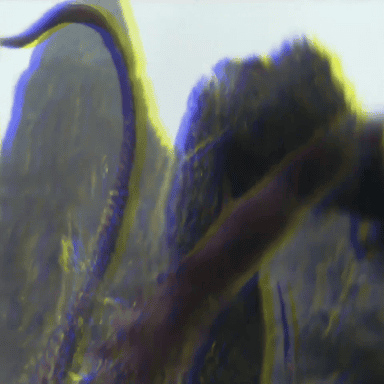} 
        &
        \includegraphics[width=.20\textwidth,height=.16\textwidth,clip,valign=c]{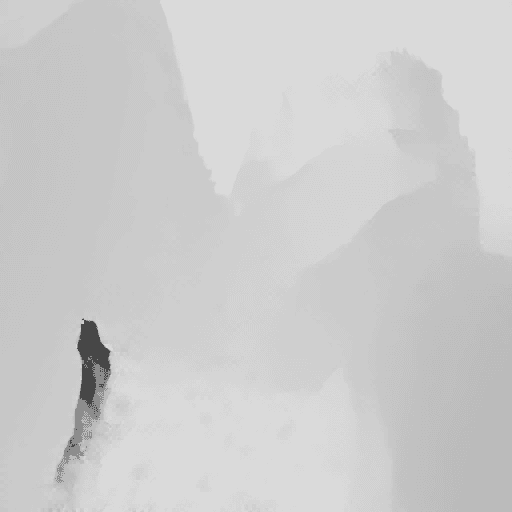} 
        &
        \includegraphics[width=.20\textwidth,height=.16\textwidth,clip,valign=c]{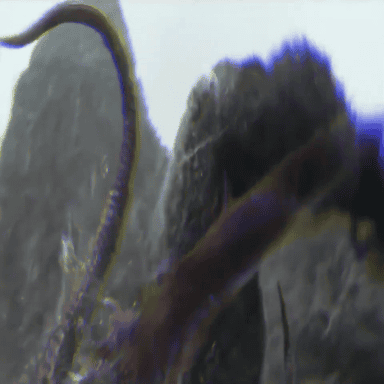} \\
        \includegraphics[width=.20\textwidth,height=.16\textwidth,clip,valign=c]{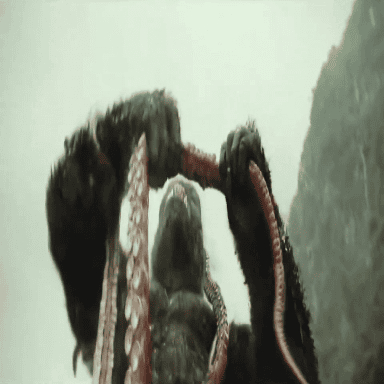}
        &\includegraphics[width=.20\textwidth,height=.16\textwidth,clip,valign=c]{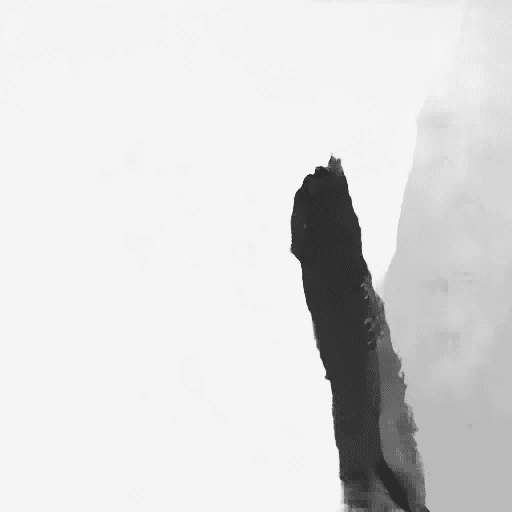}
        &
        \includegraphics[width=.20\textwidth,height=.16\textwidth,clip,valign=c]{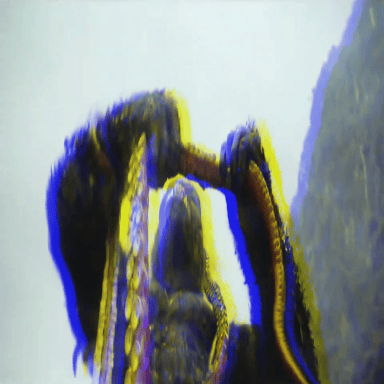} 
        &
        \includegraphics[width=.20\textwidth,height=.16\textwidth,clip,valign=c]{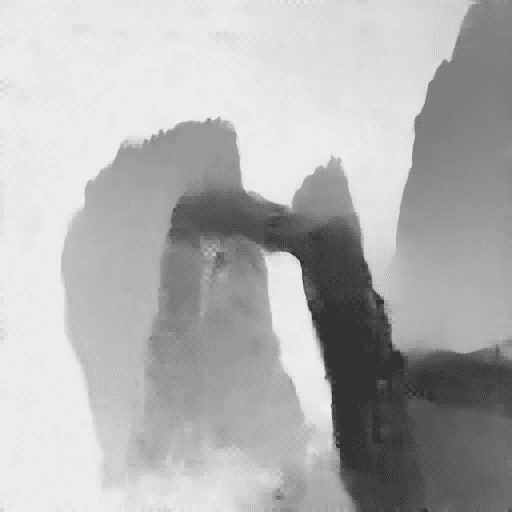} 
        &
        \includegraphics[width=.20\textwidth,height=.16\textwidth,clip,valign=c]{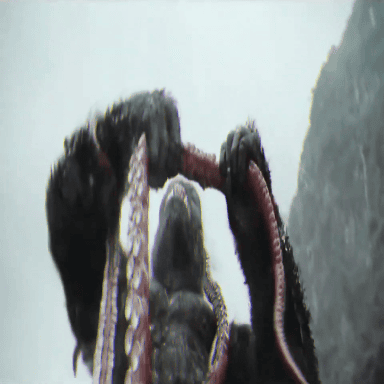} \\
        \includegraphics[width=.20\textwidth,height=.16\textwidth,clip,valign=c]{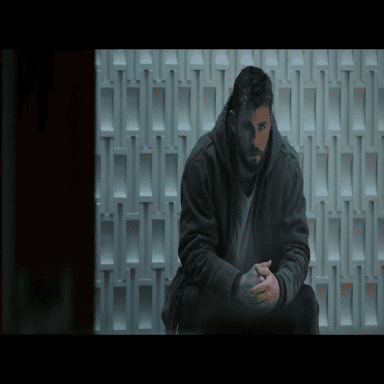}
        &\includegraphics[width=.20\textwidth,height=.16\textwidth,clip,valign=c]{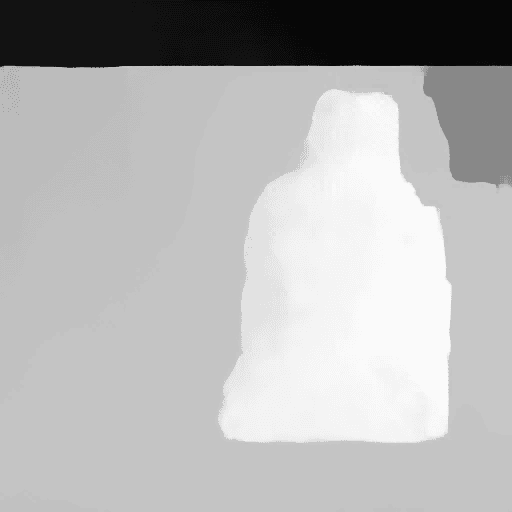}
        &
        \includegraphics[width=.20\textwidth,height=.16\textwidth,clip,valign=c]{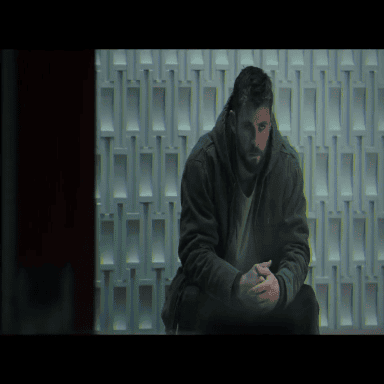} 
        &
        \includegraphics[width=.20\textwidth,height=.16\textwidth,clip,valign=c]{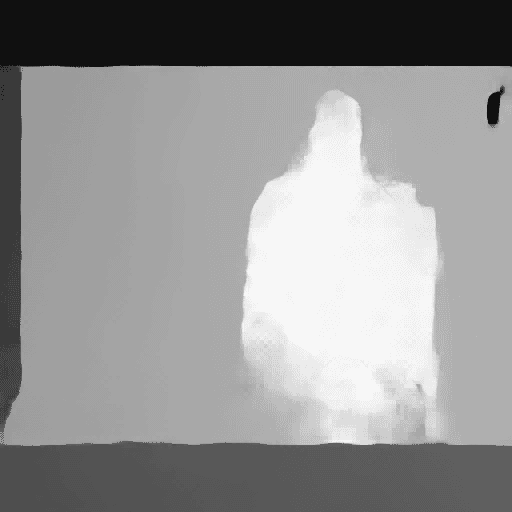} 
        &
        \includegraphics[width=.20\textwidth,height=.16\textwidth,clip,valign=c]{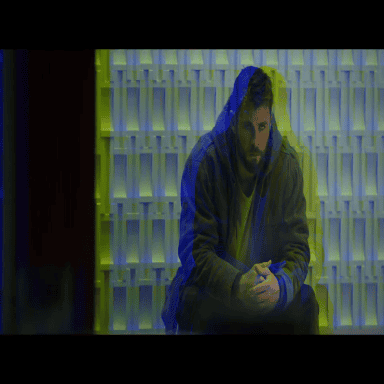} \\
        \includegraphics[width=.20\textwidth,height=.16\textwidth,clip,valign=c]{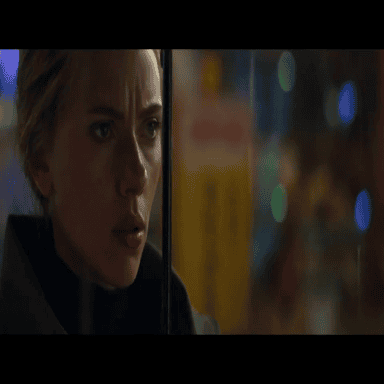}
        &\includegraphics[width=.20\textwidth,height=.16\textwidth,clip,valign=c]{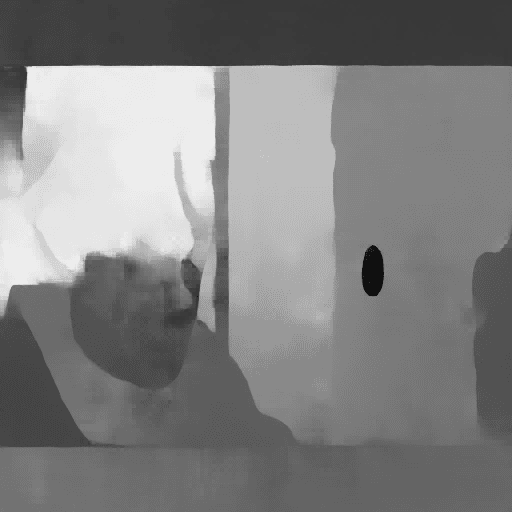}
        &
        \includegraphics[width=.20\textwidth,height=.16\textwidth,clip,valign=c]{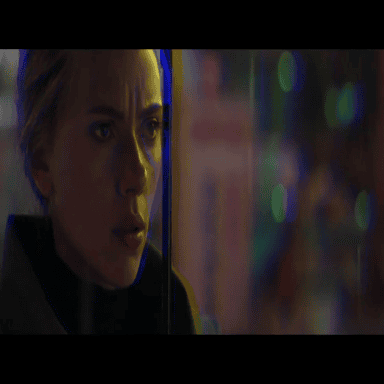} 
        &
        \includegraphics[width=.20\textwidth,height=.16\textwidth,clip,valign=c]{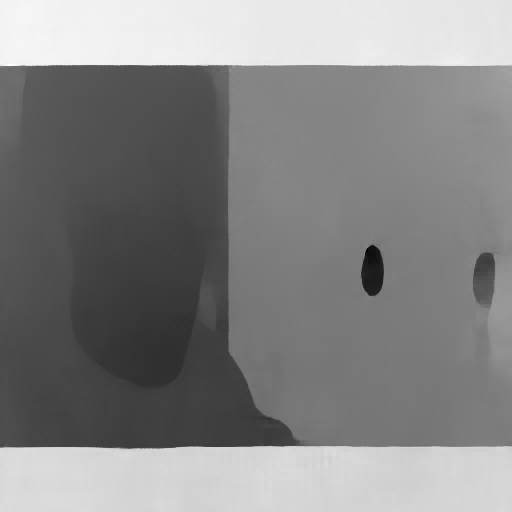} 
        &
        \includegraphics[width=.20\textwidth,height=.16\textwidth,clip,valign=c]{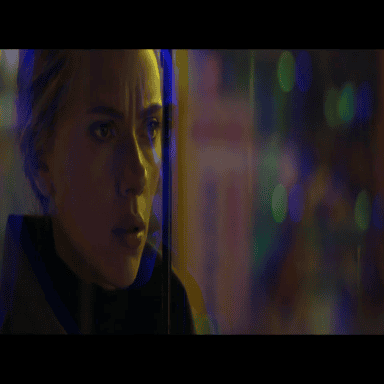} \\
    \end{tblr}
        \caption{Visual demonstration of the disparity analysis results. Our network predicts reasonable stereo effects but may be stronger or weaker if compared to the ground truth. The L2R disparity computes the left-to-right disparity using RAFT-Stereo~\citep{lipson2021raft}.}
        \label{fig:wrong_stereo}
\end{figure}

\noindent\textit{Benchmark results.} 
Our qualitative and quantitative results are shown in~\Cref{fig:teaser} and~\Cref{tab:main_table}, respectively. The visual results show that other methods tend to generate right-view images with texture deformations.
To be specific, 3D photo struggles to find accurate depth cues with \textit{MiDaS}~\citep{birkl2023midas} depth estimation model, resulting in inaccurate warping on given images. Stereo-from-mono can generate images well but often comes with unpleasant black dots around the warping shapes. 
StereoDiffusion requires using null-text inversion~\cite{mokady2023null} to convert a given image to the latent space and then warp the latent features to create the right-view image. It highly depends on the performance of the inversion, which creates unstable performances. As shown in our table, our method yields better numerical results. This finding aligns with the visual results. In addition, our accompanying videos demonstrate better stability in terms of jittering and shaking. Please watch the accompanying videos with 0.5 speed to see the artifacts generated by the different methods.

\noindent\textit{Ablation results.} \Cref{tab:main_table} shows our ablation results. We show that our method is not going to converge without using implicit disparity guidance, while a significant performance drop may occur when removing our proposed layered disparity. We show that our layered disparity generates better visual quality in~\Cref{fig:disp_diff} compared to the outputs from implicit disparities. Though not significant, the attention blocks can slightly improve the overall performance, while the context fusion module contributes significantly.
Additional experiments including alternative masking strategies, the inclusion of the context fusion module, flow-guided feature propagation, and different backbone choices are included in our supplementary material.
Lastly, we show our method may generate different levels of stereo effects in~\Cref{fig:wrong_stereo} compared to the ground truth, but this is expected due to the underdetermined nature of the problem, and we consider our solution also as reasonable.

% \section{Ablations}

% This section presents the ablation studies of our proposed method.

\section{Discussion And Limitation}

To enhance the viewing experience, films sometimes employ a stronger stereoscopic effect at the start and end, while moderating it in the middle to ensure viewer comfort~\cite{neuman2009bolt,Ranftl2022}. Thus, the stereo parameters such as focal length, are hard to retrieve even for the same film.
Theoretically, the precise reproduction of the right view is impossible without knowing the stereo parameters in advance. 
By learning through a large-scale dataset, \textit{ImmersePro} estimates its average disparity, then tries to create an average-level stereo effect for input videos rather than reproduce the precise right pair.
Therefore, as shown in~\Cref{fig:wrong_stereo}, our model may produce reasonable but ``inaccurate" stereo effects if compared with the ground truth.

A reasonable stereo conversion pipeline involves a warping-and-inpainting process, where the inpainting operation fills the black holes created by the warping operation. One sample work is stereo-from-mono~\citep{watson2020learning} that performs inpainting with a randomly sampled image from the training dataset. In a way, our method can be seen as an improvement to stereo-from-mono by intelligently selecting the correct regions for inpainting.
However, this strategy works for creating stereo movies with ``subtle" stereo effects without the need for significant inpainting.
As we observed in most 3D movie examples, very few movies contain strong stereo effects.
Notably, our method cannot produce strong stereo effects due to the limited dataset and limited inpainting capabilities. In future work, we would like to investigate how Nerf~\citep{mildenhall2021nerf}-based inpainting can be used for stereo-movie generation.

\section{Conclusion}

This work presents an end-to-end video-based stereo conversion method that generates right-view video sequences according to the input video. Our method automatically utilizes layered disparity maps on top of implicit disparities. Additionally, we propose \textit{Youtube-SBS}, a large-scale stereo dataset that is publicly available for benchmarking purposes.
Extensive qualitative and quantitative evaluations demonstrated the robustness of our approach against previous works.

% In practice, we observe that apart from horizontal warping, some stereo videos also contain vertical shifts. Our method, however, is guided by an implicit estimation of horizontal shifts, which can be one limitation of this method.

% DO NOT INCLUDE ACKNOWLEDGMENTS IN AN ANONYMOUS SUBMISSION TO SIGGRAPH 2019
%\begin{acks}
%
%The authors would like to thank Dr. Maura Turolla of Telecom
%Italia for providing specifications about the application scenario.
%
%The work is supported by the \grantsponsor{GS501100001809}{National
%  Natural Science Foundation of
%  China}{http://dx.doi.org/10.13039/501100001809} under Grant
%No.:~\grantnum{GS501100001809}{61273304\_a}
%and~\grantnum[http://www.nnsf.cn/youngscientists]{GS501100001809}{Young
%  Scientists' Support Program}.
%
%
%\end{acks}

% Bibliographys

\bibliographystyle{iclr2025_conference}
\bibliography{iclr2025_conference}

\begin{thebibliography}{36}
\providecommand{\natexlab}[1]{#1}
\providecommand{\url}[1]{\texttt{#1}}
\expandafter\ifx\csname urlstyle\endcsname\relax
  \providecommand{\doi}[1]{doi: #1}\else
  \providecommand{\doi}{doi: \begingroup \urlstyle{rm}\Url}\fi

\bibitem[Birkl et~al.(2023)Birkl, Wofk, and M{\"u}ller]{birkl2023midas}
Reiner Birkl, Diana Wofk, and Matthias M{\"u}ller.
\newblock Midas v3.1 -- a model zoo for robust monocular relative depth estimation.
\newblock \emph{arXiv preprint arXiv:2307.14460}, 2023.

\bibitem[Butler et~al.(2012)Butler, Wulff, Stanley, and Black]{Butler:ECCV:2012}
D.~J. Butler, J.~Wulff, G.~B. Stanley, and M.~J. Black.
\newblock A naturalistic open source movie for optical flow evaluation.
\newblock In {A. Fitzgibbon et al. (Eds.)} (ed.), \emph{European Conf. on Computer Vision (ECCV)}, Part IV, LNCS 7577, pp.\  611--625. Springer-Verlag, oct 2012.

\bibitem[Chen et~al.(2019)Chen, Yuan, and Bao]{Chen2019}
Bei Chen, Jiabin Yuan, and Xiuping Bao.
\newblock Automatic 2d-to-3d video conversion using 3d densely connected convolutional networks.
\newblock In \emph{2019 IEEE 31st International Conference on Tools with Artificial Intelligence (ICTAI)}. IEEE, November 2019.
\newblock \doi{10.1109/ictai.2019.00058}.
\newblock URL \url{http://dx.doi.org/10.1109/ICTAI.2019.00058}.

\bibitem[Dai et~al.(2017)Dai, Qi, Xiong, Li, Zhang, Hu, and Wei]{dai2017deformable}
Jifeng Dai, Haozhi Qi, Yuwen Xiong, Yi~Li, Guodong Zhang, Han Hu, and Yichen Wei.
\newblock Deformable convolutional networks.
\newblock In \emph{Proceedings of the IEEE international conference on computer vision}, pp.\  764--773, 2017.

\bibitem[Devernay \& Beardsley(2010)Devernay and Beardsley]{Devernay2010}
Frédéric Devernay and Paul Beardsley.
\newblock \emph{Stereoscopic Cinema}, pp.\  11–51.
\newblock Springer Berlin Heidelberg, 2010.
\newblock ISBN 9783642123924.
\newblock \doi{10.1007/978-3-642-12392-4_2}.
\newblock URL \url{http://dx.doi.org/10.1007/978-3-642-12392-4_2}.

\bibitem[Duke et~al.(2021)Duke, Ahmed, Wolf, Aarabi, and Taylor]{duke2021sstvos}
Brendan Duke, Abdalla Ahmed, Christian Wolf, Parham Aarabi, and Graham~W Taylor.
\newblock Sstvos: Sparse spatiotemporal transformers for video object segmentation.
\newblock In \emph{Proceedings of the IEEE/CVF conference on computer vision and pattern recognition}, pp.\  5912--5921, 2021.

\bibitem[Haris et~al.(2019)Haris, Shakhnarovich, and Ukita]{haris2019recurrent}
Muhammad Haris, Gregory Shakhnarovich, and Norimichi Ukita.
\newblock Recurrent back-projection network for video super-resolution.
\newblock In \emph{Proceedings of the IEEE/CVF conference on computer vision and pattern recognition}, pp.\  3897--3906, 2019.

\bibitem[Ilg et~al.(2017)Ilg, Mayer, Saikia, Keuper, Dosovitskiy, and Brox]{IMKDB17}
E.~Ilg, N.~Mayer, T.~Saikia, M.~Keuper, A.~Dosovitskiy, and T.~Brox.
\newblock Flownet 2.0: Evolution of optical flow estimation with deep networks.
\newblock In \emph{IEEE Conference on Computer Vision and Pattern Recognition (CVPR)}, Jul 2017.
\newblock URL \url{http://lmb.informatik.uni-freiburg.de//Publications/2017/IMKDB17}.

\bibitem[Jaderberg et~al.(2015)Jaderberg, Simonyan, Zisserman, et~al.]{jaderberg2015spatial}
Max Jaderberg, Karen Simonyan, Andrew Zisserman, et~al.
\newblock Spatial transformer networks.
\newblock \emph{Advances in neural information processing systems}, 28, 2015.

\bibitem[Li et~al.(2023)Li, Min, Tripathi, and Vasconcelos]{li2023svitt}
Yi~Li, Kyle Min, Subarna Tripathi, and Nuno Vasconcelos.
\newblock Svitt: Temporal learning of sparse video-text transformers.
\newblock In \emph{Proceedings of the IEEE/CVF Conference on Computer Vision and Pattern Recognition}, pp.\  18919--18929, 2023.

\bibitem[Li et~al.(2021)Li, Liu, Drenkow, Ding, Creighton, Taylor, and Unberath]{li2021revisiting}
Zhaoshuo Li, Xingtong Liu, Nathan Drenkow, Andy Ding, Francis~X Creighton, Russell~H Taylor, and Mathias Unberath.
\newblock Revisiting stereo depth estimation from a sequence-to-sequence perspective with transformers.
\newblock In \emph{Proceedings of the IEEE/CVF international conference on computer vision}, pp.\  6197--6206, 2021.

\bibitem[Li et~al.(2022)Li, Lu, Qin, Guo, and Cheng]{liCvpr22vInpainting}
Zhen Li, Cheng-Ze Lu, Jianhua Qin, Chun-Le Guo, and Ming-Ming Cheng.
\newblock Towards an end-to-end framework for flow-guided video inpainting.
\newblock In \emph{IEEE Conference on Computer Vision and Pattern Recognition (CVPR)}, 2022.

\bibitem[Lipson et~al.(2021)Lipson, Teed, and Deng]{lipson2021raft}
Lahav Lipson, Zachary Teed, and Jia Deng.
\newblock Raft-stereo: Multilevel recurrent field transforms for stereo matching.
\newblock In \emph{International Conference on 3D Vision (3DV)}, 2021.

\bibitem[Liu et~al.(2021)Liu, Deng, Huang, Shi, Lu, Sun, Wang, Dai, and Li]{liu2021fuseformer}
Rui Liu, Hanming Deng, Yangyi Huang, Xiaoyu Shi, Lewei Lu, Wenxiu Sun, Xiaogang Wang, Jifeng Dai, and Hongsheng Li.
\newblock Fuseformer: Fusing fine-grained information in transformers for video inpainting.
\newblock In \emph{Proceedings of the IEEE/CVF international conference on computer vision}, pp.\  14040--14049, 2021.

\bibitem[Loshchilov \& Hutter(2017)Loshchilov and Hutter]{loshchilov2017decoupled}
Ilya Loshchilov and Frank Hutter.
\newblock Decoupled weight decay regularization.
\newblock \emph{arXiv preprint arXiv:1711.05101}, 2017.

\bibitem[Mehl et~al.(2024)Mehl, Bruhn, Gross, and Schroers]{mehl2024stereo}
Lukas Mehl, Andr{\'e}s Bruhn, Markus Gross, and Christopher Schroers.
\newblock Stereo conversion with disparity-aware warping, compositing and inpainting.
\newblock In \emph{Proceedings of the IEEE/CVF Winter Conference on Applications of Computer Vision}, pp.\  4260--4269, 2024.

\bibitem[Menze \& Geiger(2015)Menze and Geiger]{Menze2015CVPR}
Moritz Menze and Andreas Geiger.
\newblock Object scene flow for autonomous vehicles.
\newblock In \emph{Conference on Computer Vision and Pattern Recognition (CVPR)}, 2015.

\bibitem[Mildenhall et~al.(2021)Mildenhall, Srinivasan, Tancik, Barron, Ramamoorthi, and Ng]{mildenhall2021nerf}
Ben Mildenhall, Pratul~P Srinivasan, Matthew Tancik, Jonathan~T Barron, Ravi Ramamoorthi, and Ren Ng.
\newblock Nerf: Representing scenes as neural radiance fields for view synthesis.
\newblock \emph{Communications of the ACM}, 65\penalty0 (1):\penalty0 99--106, 2021.

\bibitem[Mokady et~al.(2023)Mokady, Hertz, Aberman, Pritch, and Cohen-Or]{mokady2023null}
Ron Mokady, Amir Hertz, Kfir Aberman, Yael Pritch, and Daniel Cohen-Or.
\newblock Null-text inversion for editing real images using guided diffusion models.
\newblock In \emph{Proceedings of the IEEE/CVF Conference on Computer Vision and Pattern Recognition}, pp.\  6038--6047, 2023.

\bibitem[Neuman(2009)]{neuman2009bolt}
Robert Neuman.
\newblock Bolt 3d: a case study.
\newblock In \emph{Stereoscopic Displays and Applications XX}, volume 7237, pp.\  133--142. SPIE, 2009.

\bibitem[Ranftl et~al.(2022)Ranftl, Lasinger, Hafner, Schindler, and Koltun]{Ranftl2022}
Rene Ranftl, Katrin Lasinger, David Hafner, Konrad Schindler, and Vladlen Koltun.
\newblock Towards robust monocular depth estimation: Mixing datasets for zero-shot cross-dataset transfer.
\newblock \emph{IEEE Transactions on Pattern Analysis and Machine Intelligence}, 44\penalty0 (3):\penalty0 1623–1637, March 2022.
\newblock ISSN 1939-3539.
\newblock \doi{10.1109/tpami.2020.3019967}.
\newblock URL \url{http://dx.doi.org/10.1109/TPAMI.2020.3019967}.

\bibitem[Shih et~al.(2020)Shih, Su, Kopf, and Huang]{Shih3DP20}
Meng-Li Shih, Shih-Yang Su, Johannes Kopf, and Jia-Bin Huang.
\newblock 3d photography using context-aware layered depth inpainting.
\newblock In \emph{IEEE Conference on Computer Vision and Pattern Recognition (CVPR)}, 2020.

\bibitem[Teed \& Deng(2020)Teed and Deng]{teed2020raft}
Zachary Teed and Jia Deng.
\newblock Raft: Recurrent all-pairs field transforms for optical flow.
\newblock In \emph{Computer Vision--ECCV 2020: 16th European Conference, Glasgow, UK, August 23--28, 2020, Proceedings, Part II 16}, pp.\  402--419. Springer, 2020.

\bibitem[Wang et~al.(2019{\natexlab{a}})Wang, Lucey, Perazzi, and Wang]{wang2019web}
Chaoyang Wang, Simon Lucey, Federico Perazzi, and Oliver Wang.
\newblock Web stereo video supervision for depth prediction from dynamic scenes.
\newblock In \emph{2019 International Conference on 3D Vision (3DV)}, pp.\  348--357. IEEE, 2019{\natexlab{a}}.

\bibitem[Wang et~al.(2024)Wang, Frisvad, Jensen, and Bigdeli]{wang2024stereodiffusion}
Lezhong Wang, Jeppe~Revall Frisvad, Mark~Bo Jensen, and Siavash~Arjomand Bigdeli.
\newblock Stereodiffusion: Training-free stereo image generation using latent diffusion models.
\newblock \emph{arXiv preprint arXiv:2403.04965}, 2024.

\bibitem[Wang et~al.(2019{\natexlab{b}})Wang, Chan, Yu, Dong, and Change~Loy]{wang2019edvr}
Xintao Wang, Kelvin~CK Chan, Ke~Yu, Chao Dong, and Chen Change~Loy.
\newblock Edvr: Video restoration with enhanced deformable convolutional networks.
\newblock In \emph{Proceedings of the IEEE/CVF conference on computer vision and pattern recognition workshops}, pp.\  0--0, 2019{\natexlab{b}}.

\bibitem[Wang et~al.(2023)Wang, Shi, Li, Huang, Cao, Zhang, Xian, and Lin]{wang2023neural}
Yiran Wang, Min Shi, Jiaqi Li, Zihao Huang, Zhiguo Cao, Jianming Zhang, Ke~Xian, and Guosheng Lin.
\newblock Neural video depth stabilizer.
\newblock \emph{arXiv preprint arXiv:2307.08695}, 2023.

\bibitem[Watson et~al.(2020)Watson, Aodha, Turmukhambetov, Brostow, and Firman]{watson2020learning}
Jamie Watson, Oisin~Mac Aodha, Daniyar Turmukhambetov, Gabriel~J Brostow, and Michael Firman.
\newblock Learning stereo from single images.
\newblock In \emph{Computer Vision--ECCV 2020: 16th European Conference, Glasgow, UK, August 23--28, 2020, Proceedings, Part I 16}, pp.\  722--740. Springer, 2020.

\bibitem[Xie et~al.(2016)Xie, Girshick, and Farhadi]{xie2016deep3d}
Junyuan Xie, Ross Girshick, and Ali Farhadi.
\newblock Deep3d: Fully automatic 2d-to-3d video conversion with deep convolutional neural networks.
\newblock In \emph{Computer Vision--ECCV 2016: 14th European Conference, Amsterdam, The Netherlands, October 11--14, 2016, Proceedings, Part IV 14}, pp.\  842--857. Springer, 2016.

\bibitem[Xue et~al.(2019)Xue, Chen, Wu, Wei, and Freeman]{xue2019video}
Tianfan Xue, Baian Chen, Jiajun Wu, Donglai Wei, and William~T Freeman.
\newblock Video enhancement with task-oriented flow.
\newblock \emph{International Journal of Computer Vision (IJCV)}, 127\penalty0 (8):\penalty0 1106--1125, 2019.

\bibitem[Yang et~al.(2024)Yang, Kang, Huang, Xu, Feng, and Zhao]{yang2024depth}
Lihe Yang, Bingyi Kang, Zilong Huang, Xiaogang Xu, Jiashi Feng, and Hengshuang Zhao.
\newblock Depth anything: Unleashing the power of large-scale unlabeled data.
\newblock \emph{arXiv preprint arXiv:2401.10891}, 2024.

\bibitem[Zhang et~al.(2018)Zhang, Isola, Efros, Shechtman, and Wang]{zhang2018perceptual}
Richard Zhang, Phillip Isola, Alexei~A Efros, Eli Shechtman, and Oliver Wang.
\newblock The unreasonable effectiveness of deep features as a perceptual metric.
\newblock In \emph{CVPR}, 2018.

\bibitem[Zhang et~al.(2019)Zhang, Zou, Ren, Jiang, and Chen]{zhang2019structure}
Yu~Zhang, Dongqing Zou, Jimmy~S Ren, Zhe Jiang, and Xiaohao Chen.
\newblock Structure-preserving stereoscopic view synthesis with multi-scale adversarial correlation matching.
\newblock In \emph{Proceedings of the IEEE/CVF Conference on Computer Vision and Pattern Recognition}, pp.\  5860--5869, 2019.

\bibitem[Zhang \& Wang(2022)Zhang and Wang]{zhang2022temporal3d}
Zheyu Zhang and Ronggang Wang.
\newblock Temporal3d: 2d-to-3d video conversion network with multi-frame fusion.
\newblock In \emph{2022 4th International Conference on Advances in Computer Technology, Information Science and Communications (CTISC)}, pp.\  1--5. IEEE, 2022.

\bibitem[Zhou et~al.(2023)Zhou, Li, Chan, and Loy]{zhou2023propainter}
Shangchen Zhou, Chongyi Li, Kelvin~C.K Chan, and Chen~Change Loy.
\newblock {ProPainter}: Improving propagation and transformer for video inpainting.
\newblock In \emph{Proceedings of IEEE International Conference on Computer Vision (ICCV)}, 2023.

\bibitem[Zhu et~al.(2019)Zhu, Hu, Lin, and Dai]{zhu2019deformable}
Xizhou Zhu, Han Hu, Stephen Lin, and Jifeng Dai.
\newblock Deformable convnets v2: More deformable, better results.
\newblock In \emph{Proceedings of the IEEE/CVF conference on computer vision and pattern recognition}, pp.\  9308--9316, 2019.

\end{thebibliography}

% Appendix
\clearpage
\appendix

\section*{SUPPLEMENTARY MATERIAL}

We present implementation details and additional experiments in our supplementary material. Please watch the accompanying videos with 0.5 speed to see the artifacts generated by the different methods.

\section{Technical Details}

\subsection{Visual Reference for Stereoeffects}

We provide a visual reference for the optical flow analysis as below:

\begin{figure*}[h]
     \centering
    
     \begin{tblr}{colspec={Q[c]Q[c]Q[c]Q[c]Q[c]Q[c]}, colsep = {.1pt}, rowsep = {0.1pt}}
        & Left & Right & $F_{l\rightarrow r}$ & $F_{r\rightarrow l}$ & Occlusion
        \\
        \adjustbox{valign=m}{\rot{\Tr{\makecell{$1-\mathcal{E}=0.00$}}}}
            & 
         \includegraphics[width=.195\textwidth,height=.22\textwidth, valign=c]{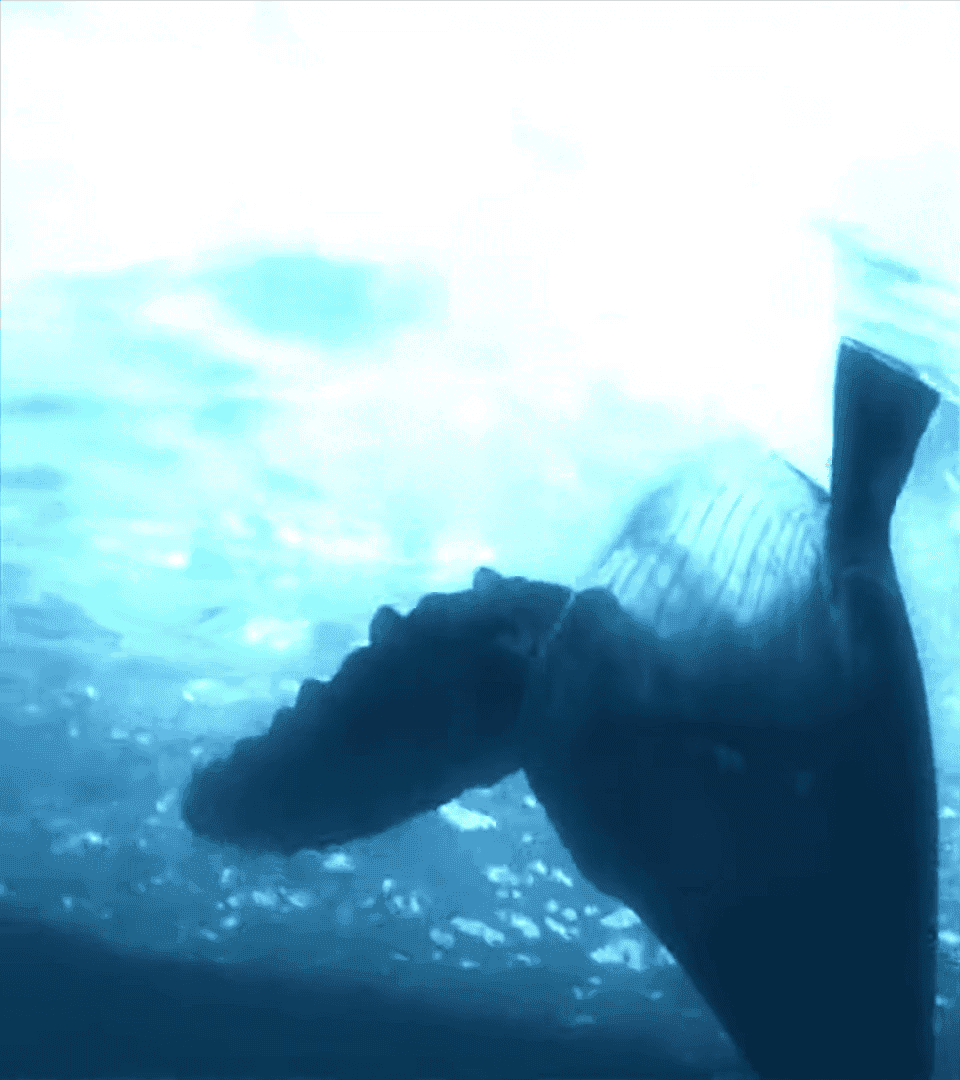}
         &
         \includegraphics[width=.195\textwidth,height=.22\textwidth, valign=c]{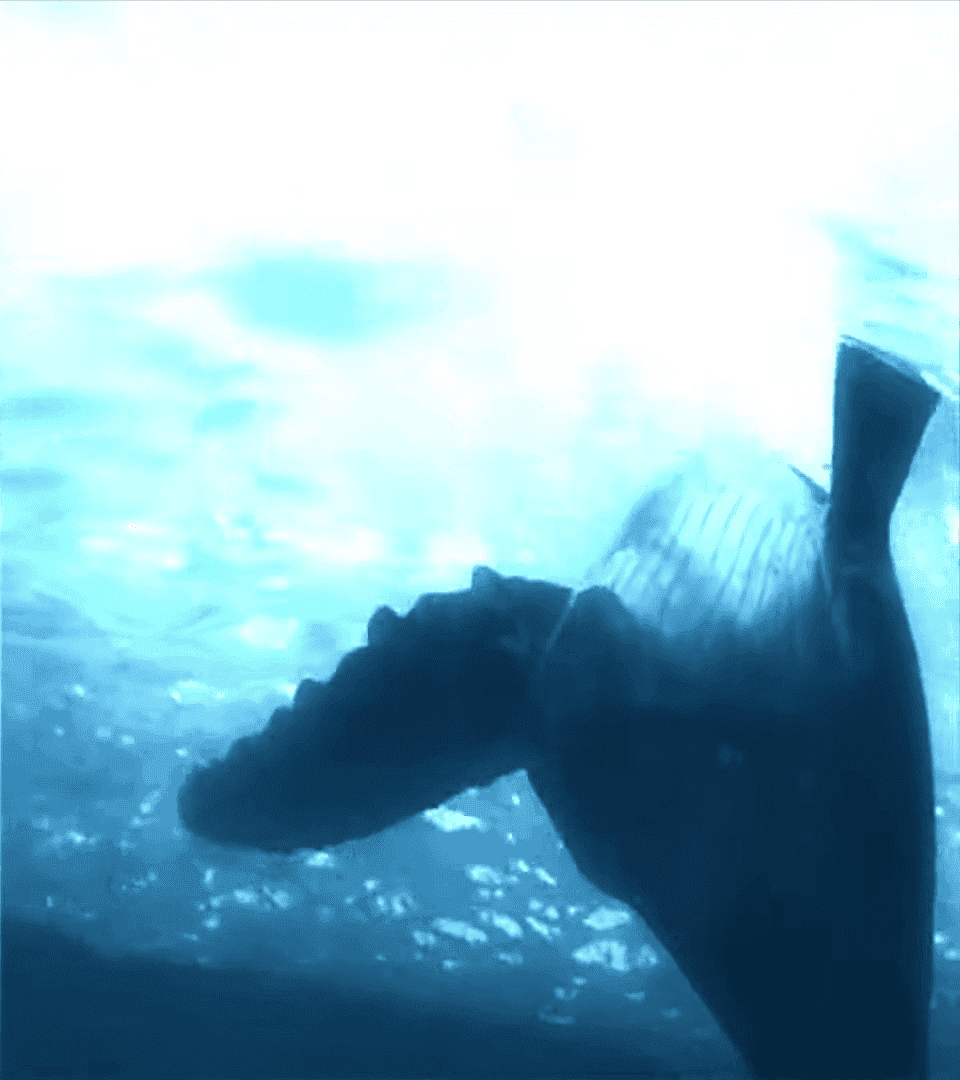}
         &
         \includegraphics[width=.195\textwidth,height=.22\textwidth, valign=c]{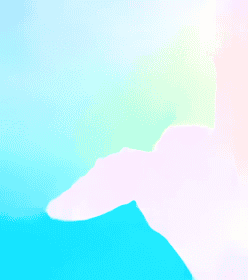}
         &
         \includegraphics[width=.195\textwidth,height=.22\textwidth, valign=c]{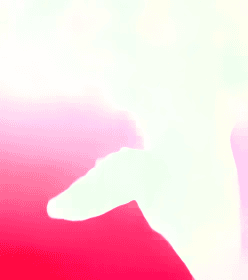}
         &
         \includegraphics[width=.195\textwidth,height=.22\textwidth, valign=c]{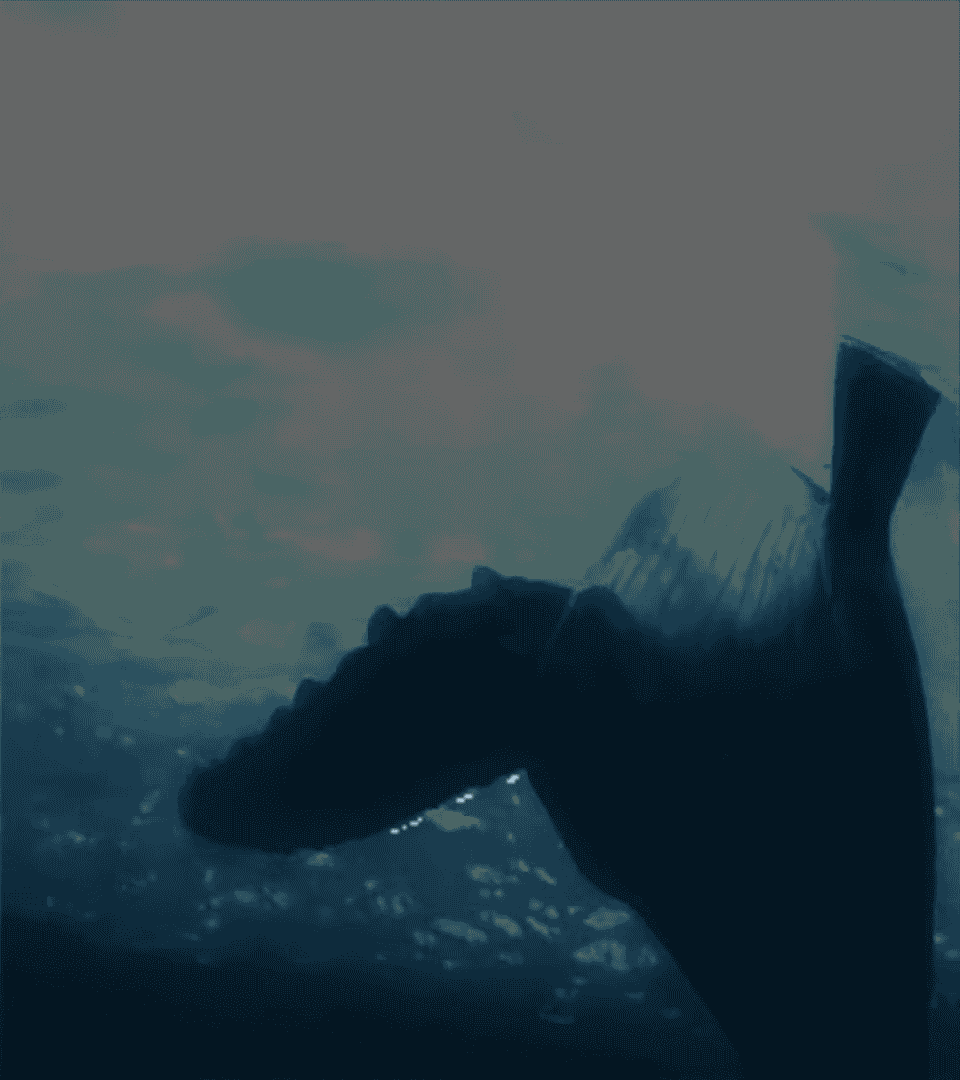}\\
        
        \adjustbox{valign=m}{\rot{\Tr{\makecell{$1-\mathcal{E}=0.05$}}}}
            & 
         \includegraphics[width=.195\textwidth,height=.22\textwidth, valign=c]{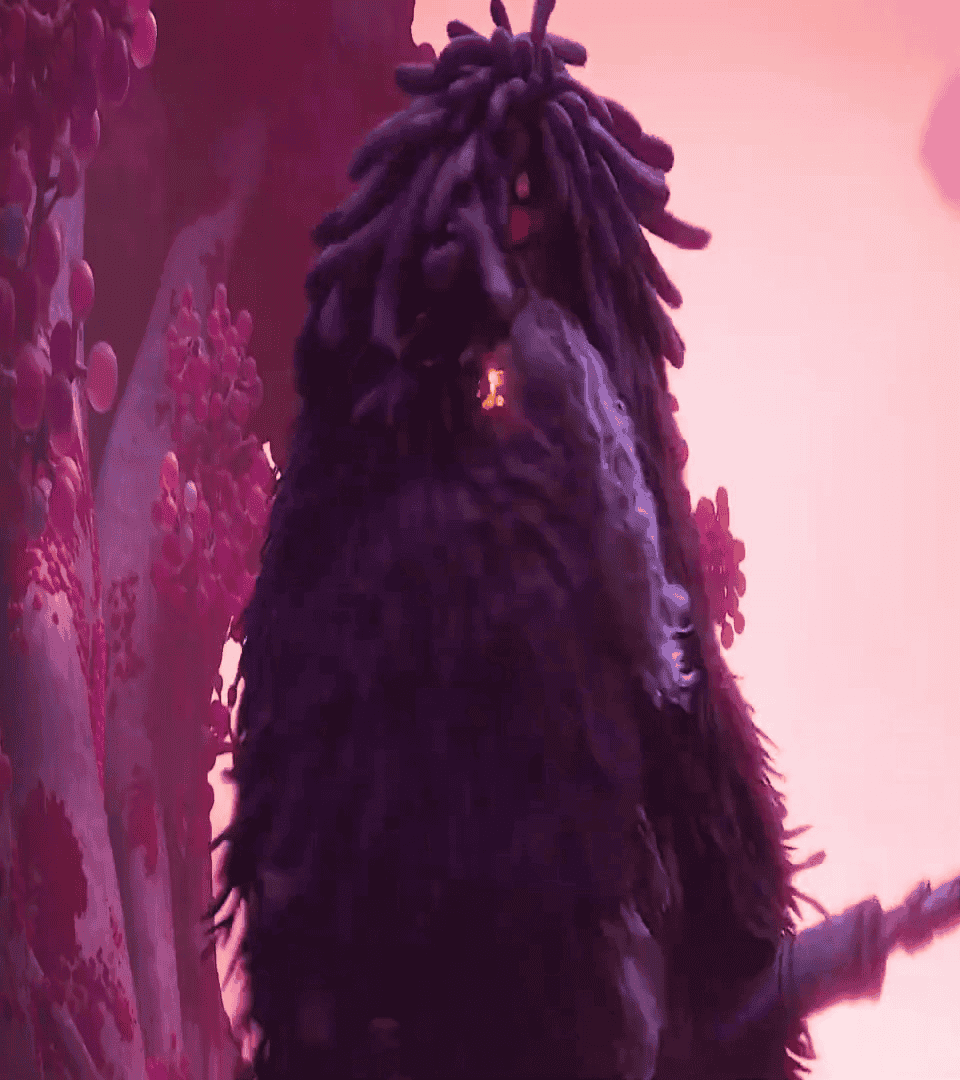}
         &
         \includegraphics[width=.195\textwidth,height=.22\textwidth, valign=c]{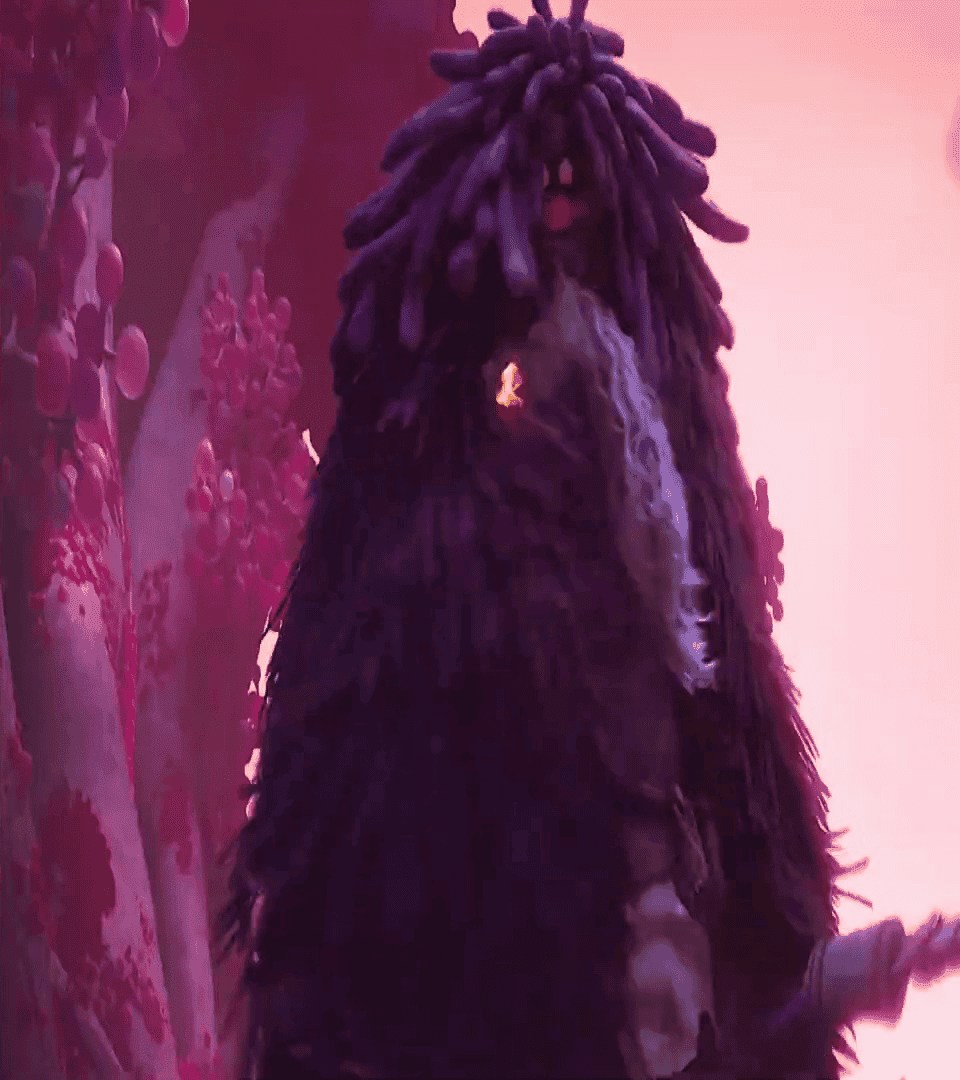}
         &
         \includegraphics[width=.195\textwidth,height=.22\textwidth, valign=c]{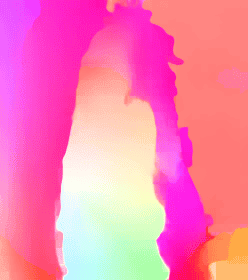}
         &
         \includegraphics[width=.195\textwidth,height=.22\textwidth, valign=c]{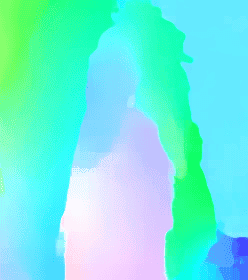}
         &
         \includegraphics[width=.195\textwidth,height=.22\textwidth, valign=c]{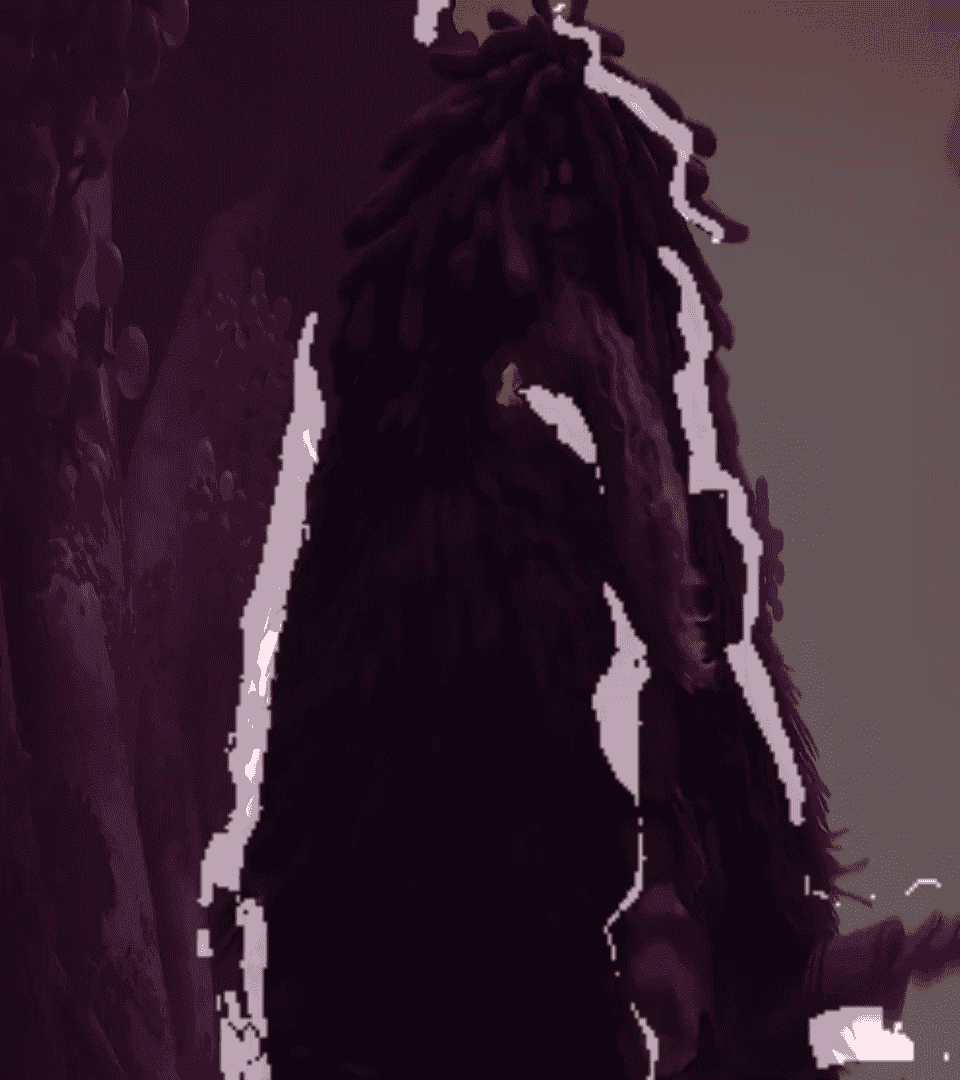}\\
        
        \adjustbox{valign=m}{\rot{\Tr{\makecell{$1-\mathcal{E}=0.12$}}}}
            & 
         \includegraphics[width=.195\textwidth,height=.22\textwidth, valign=c]{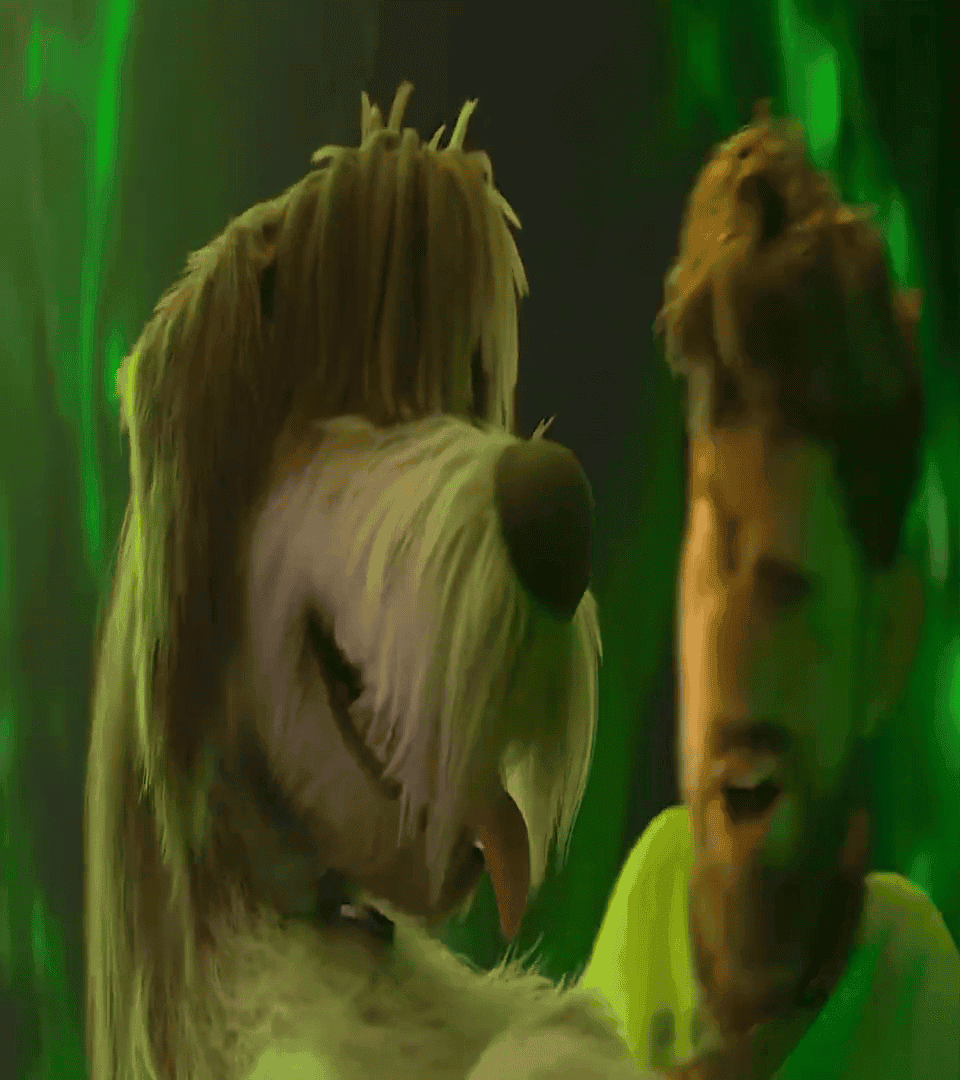}
         &
         \includegraphics[width=.195\textwidth,height=.22\textwidth, valign=c]{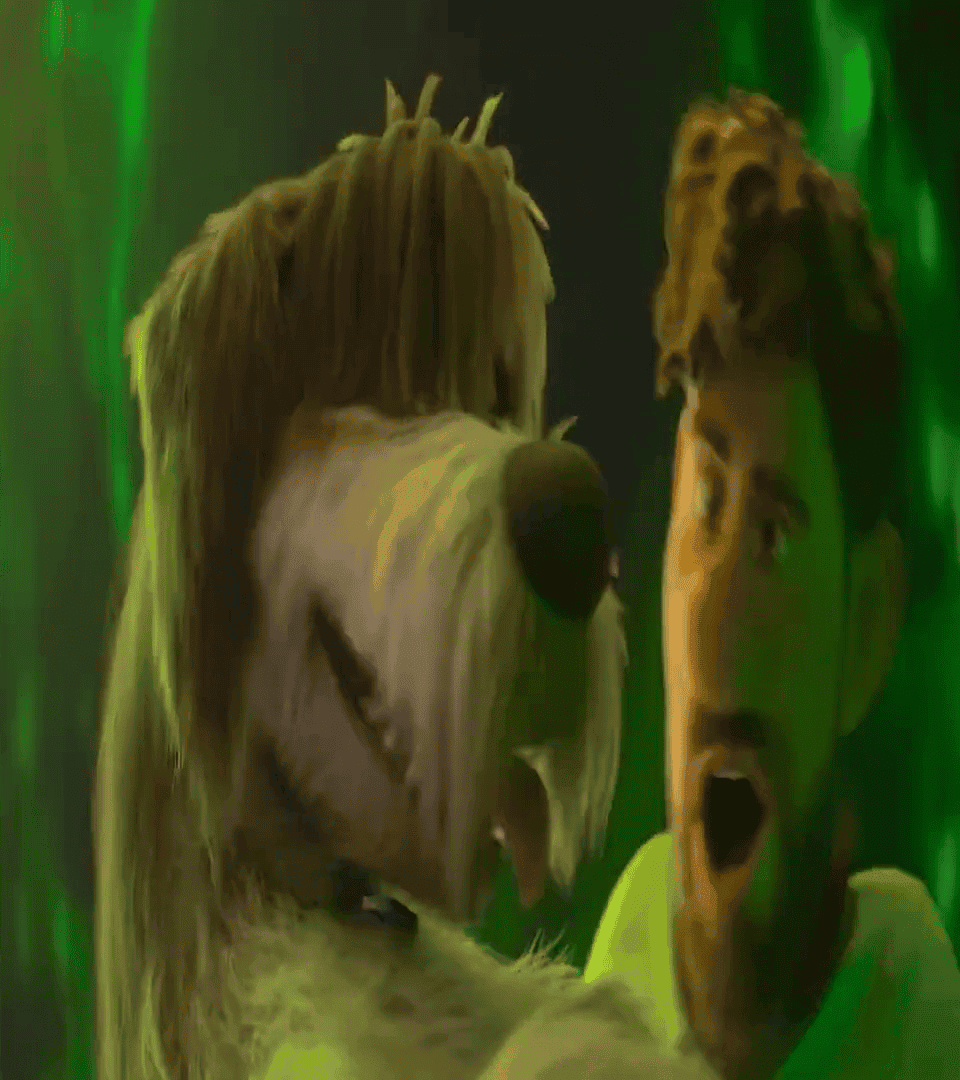}
         &
         \includegraphics[width=.195\textwidth,height=.22\textwidth, valign=c]{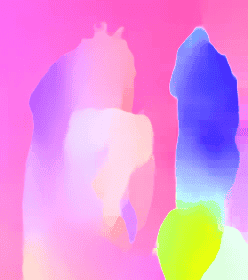}
         &
         \includegraphics[width=.195\textwidth,height=.22\textwidth, valign=c]{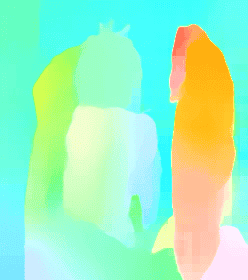}
         &
         \includegraphics[width=.195\textwidth,height=.22\textwidth, valign=c]{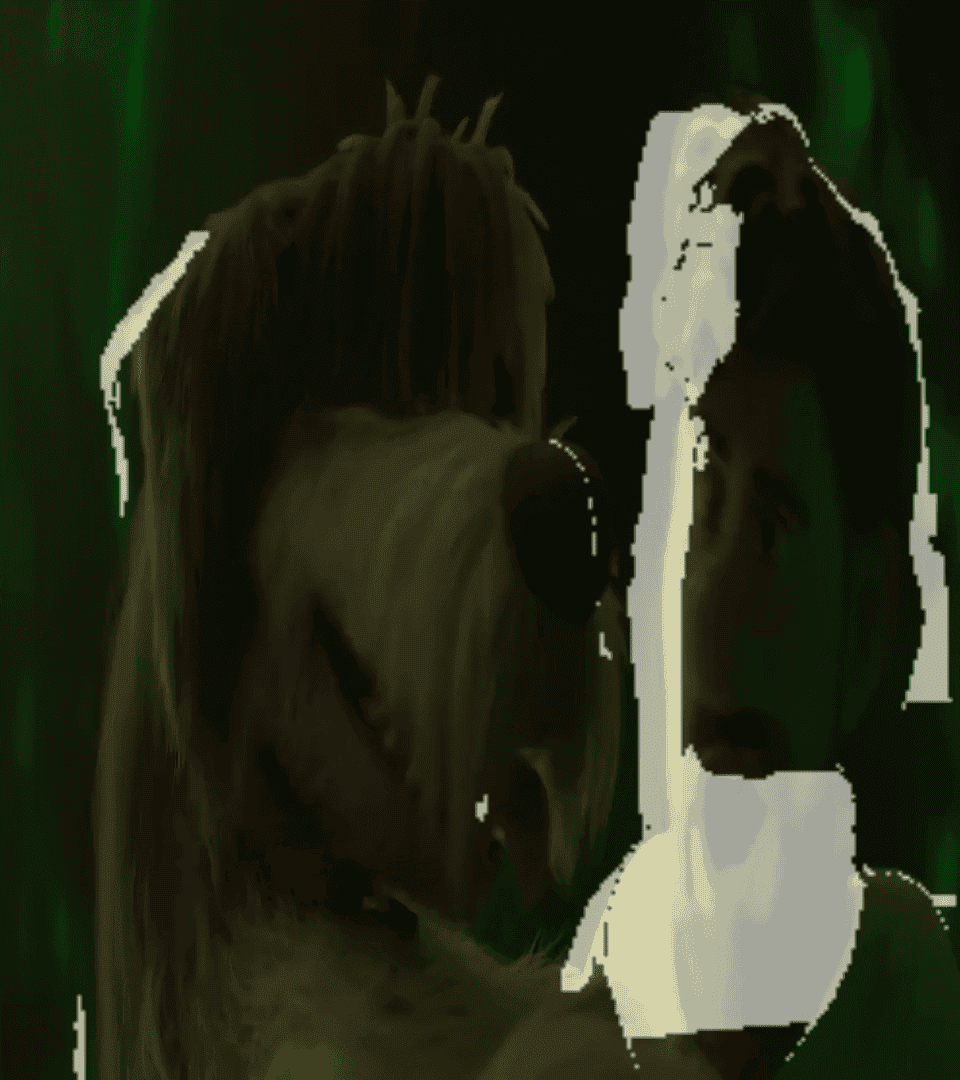}\\
        
        \adjustbox{valign=m}{\rot{\Tr{\makecell{$1-\sum\mathcal{E}=0.29$}}}}
            & 
         \includegraphics[width=.195\textwidth,height=.22\textwidth, valign=c]{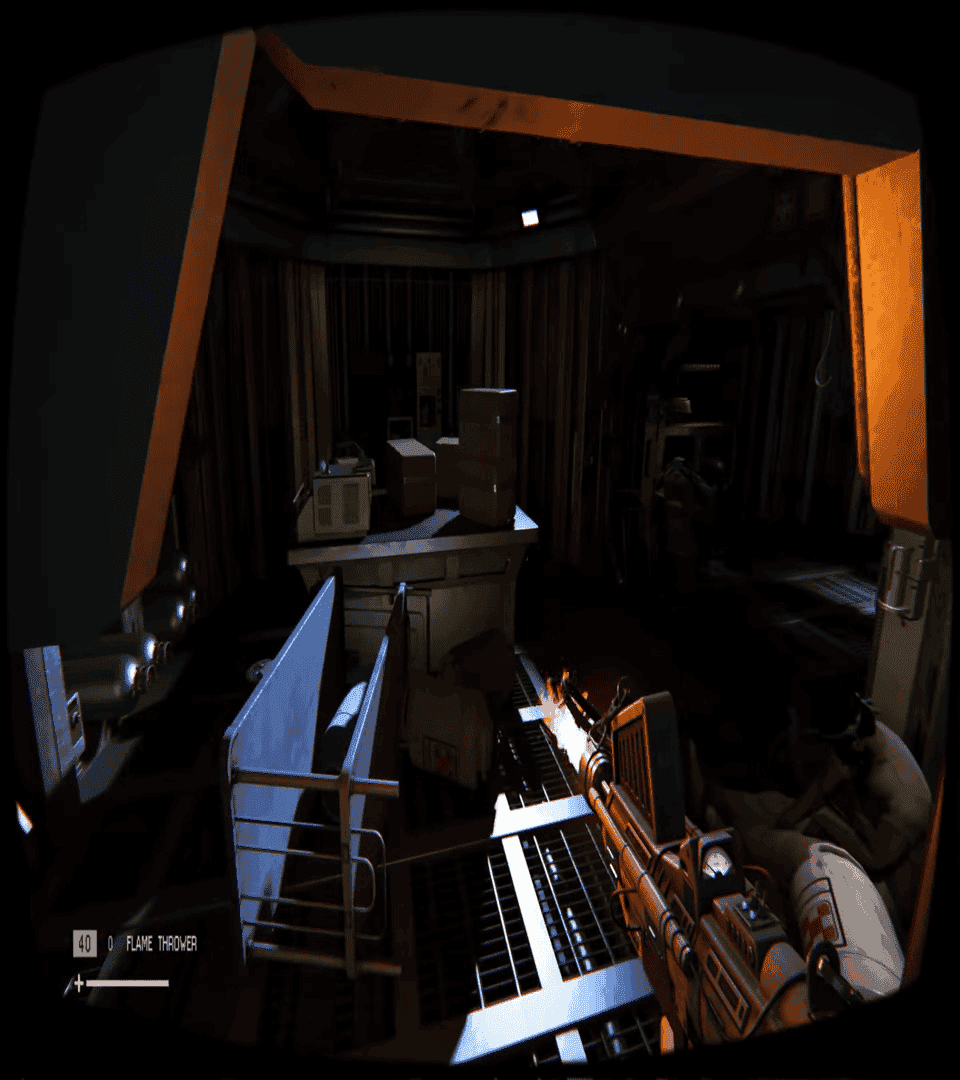}
         &
         \includegraphics[width=.195\textwidth,height=.22\textwidth, valign=c]{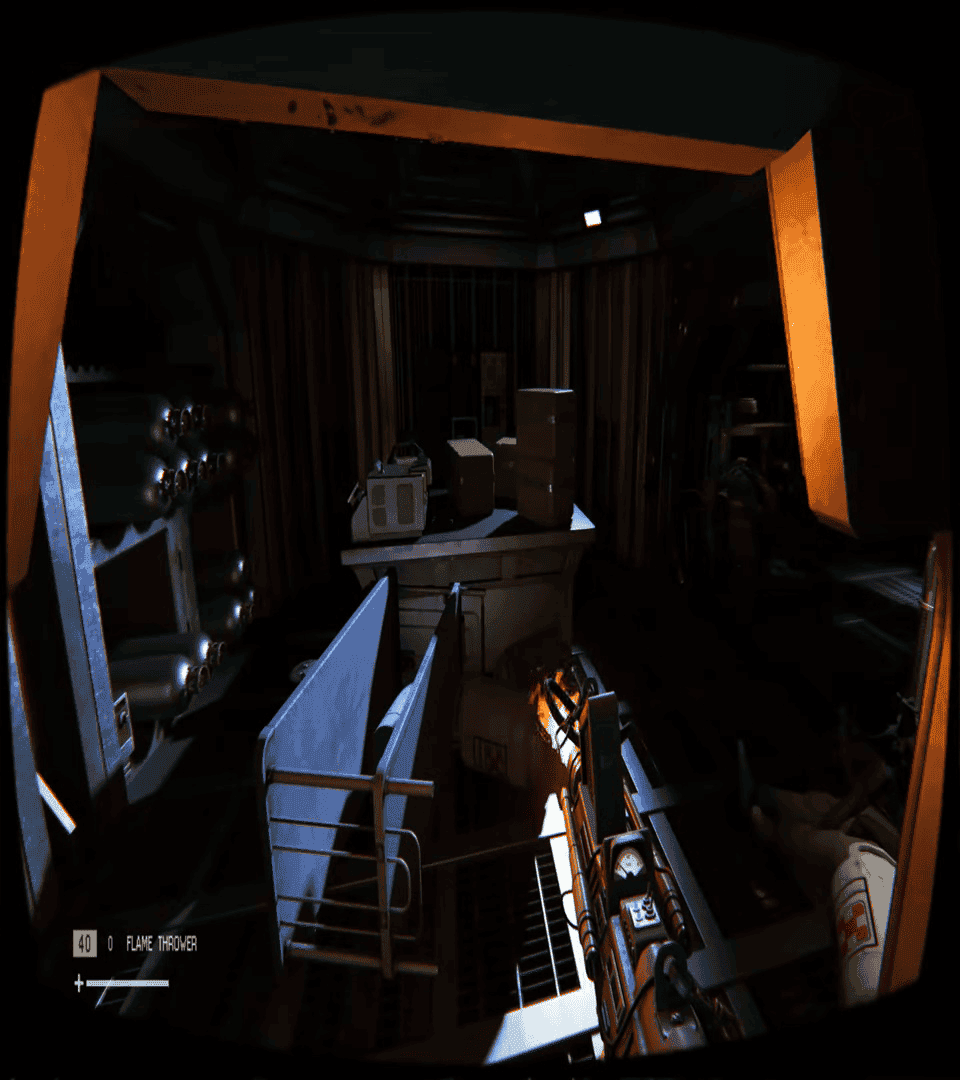}
         &
         \includegraphics[width=.195\textwidth,height=.22\textwidth, valign=c]{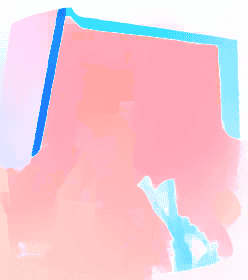}
         &
         \includegraphics[width=.195\textwidth,height=.22\textwidth, valign=c]{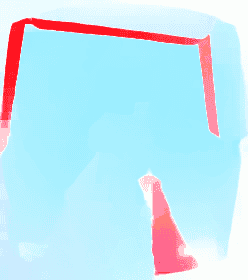}
         &
         \includegraphics[width=.195\textwidth,height=.22\textwidth, valign=c]{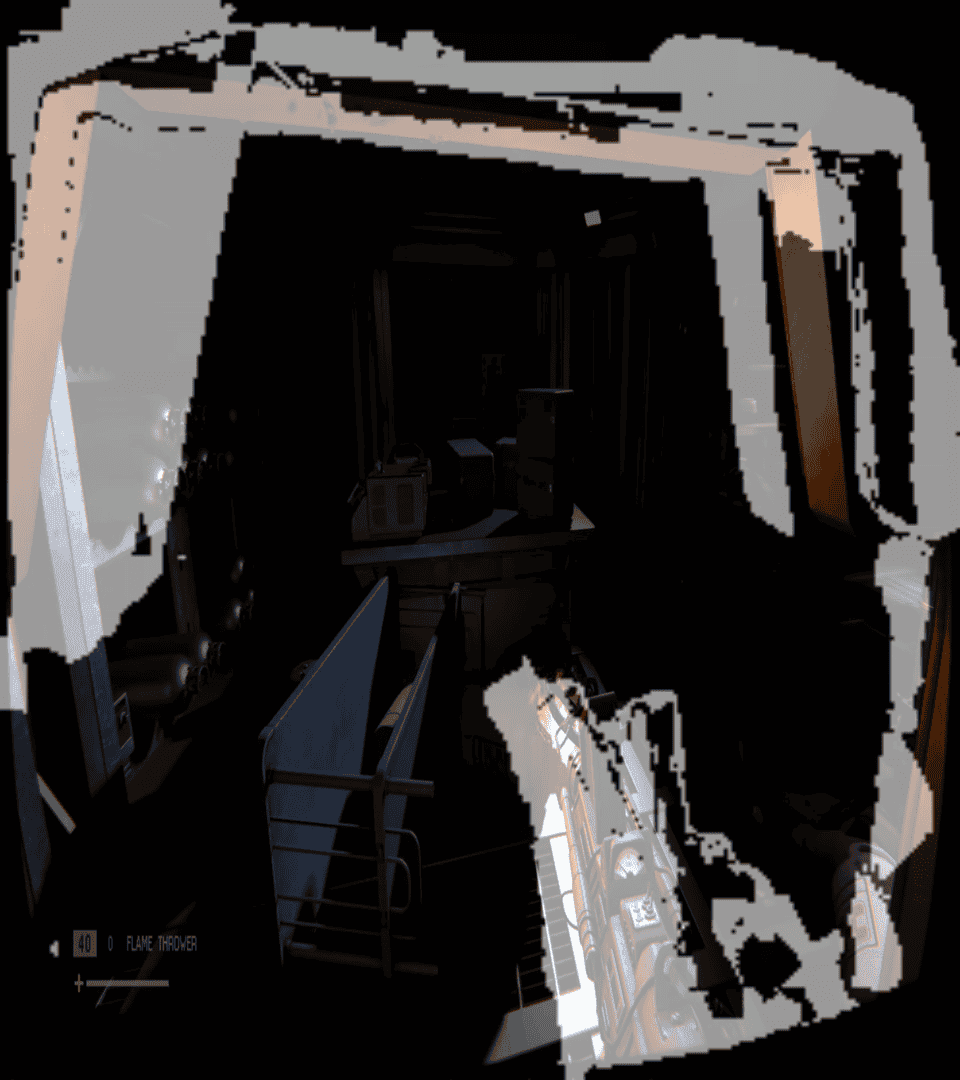}\\
        
        \adjustbox{valign=m}{\rot{\Tr{\makecell{$1-\mathcal{E}=0.59$}}}}
            & 
         \includegraphics[width=.195\textwidth,height=.22\textwidth, valign=c]{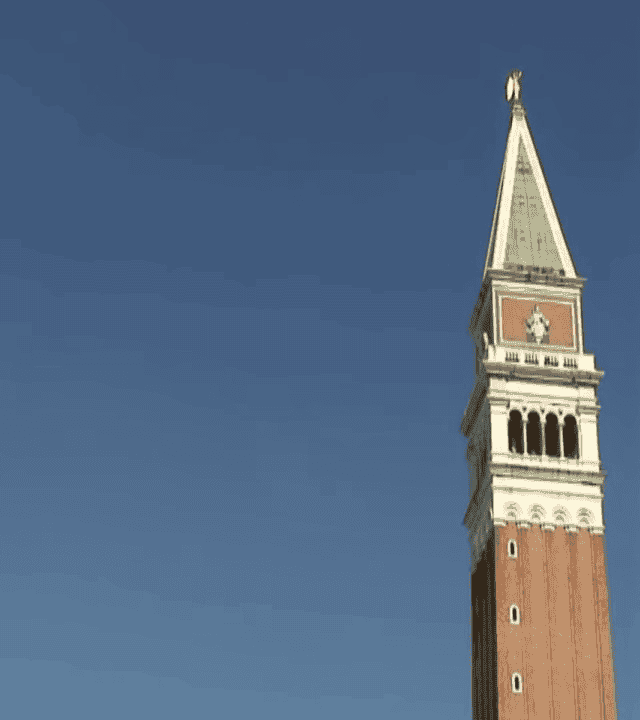}
         &
         \includegraphics[width=.195\textwidth,height=.22\textwidth, valign=c]{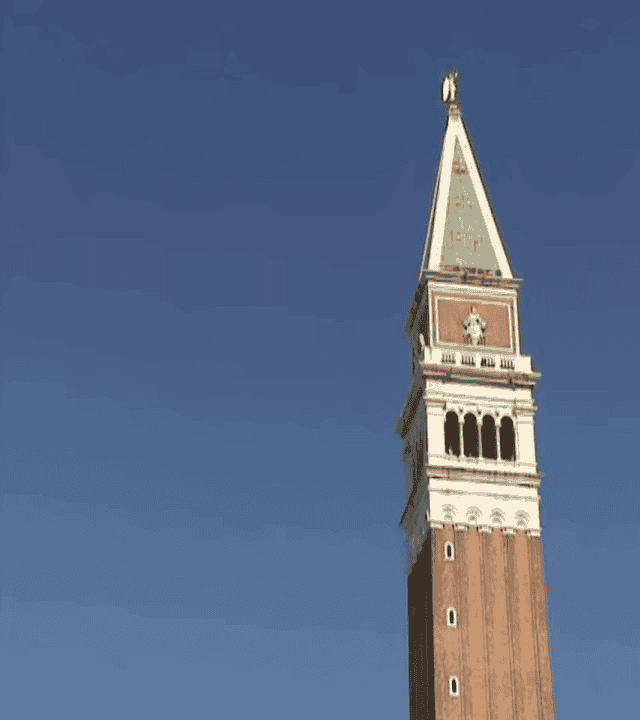}
         &
         \includegraphics[width=.195\textwidth,height=.22\textwidth, valign=c]{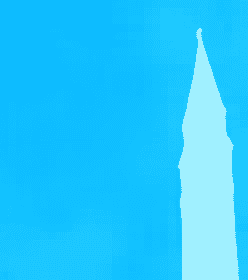}
         &
         \includegraphics[width=.195\textwidth,height=.22\textwidth, valign=c]{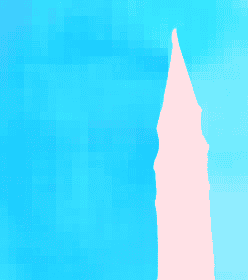}
         &
         \includegraphics[width=.195\textwidth,height=.22\textwidth, valign=c]{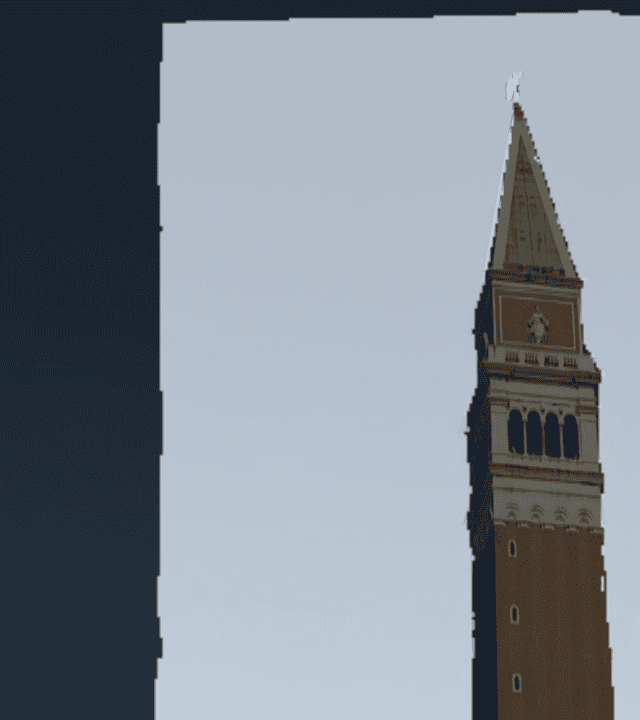}\\
    \end{tblr}

        \caption{Visual demonstration of the different levels of stereo effects with the dataset.}
        \label{fig:dataset}
\end{figure*}

\subsection{Context Encoder}
\label{sec:context_encoder}
Our context encoder uses a stack of convolutional layers to obtain semantic features from the input image as below. Starting from the $5^{th}$ convolution layer,  the extracted features are the concatenation of the features from the current and previous layers.
{\footnotesize
\begin{lstlisting}[language=Python]
Conv2d(3, 64, kernel_size=3, stride=1, padding=1),
LeakyReLU(0.2, inplace=True),
Conv2d(64, 64, kernel_size=3, stride=2, padding=1),
LeakyReLU(0.2, inplace=True),
Conv2d(64, 128, kernel_size=3, stride=1, padding=1),
LeakyReLU(0.2, inplace=True),
Conv2d(128, 256, kernel_size=3, stride=1, padding=1),
LeakyReLU(0.2, inplace=True),
Conv2d(256,384,kernel_size=3,stride=1,padding=1,groups=1),
LeakyReLU(0.2, inplace=True),
Conv2d(640,512,kernel_size=3,stride=1,padding=1,groups=2),
LeakyReLU(0.2, inplace=True),
Conv2d(768,384,kernel_size=3,stride=1,padding=1,groups=4),
LeakyReLU(0.2, inplace=True),
Conv2d(640,256,kernel_size=3,stride=1,padding=1,groups=8),
LeakyReLU(0.2, inplace=True),
\end{lstlisting}
}

\section{Additional Experiments}

\subsection{Alternative Mask Selection Algorithm}
\label{sec:algo_alt}

\begin{figure}[h]
     \centering

     \begin{tblr}{colspec={Q[c]Q[c]Q[c]Q[c]}, colsep = {.3pt}, rowsep = {.3pt}}
        \scriptsize Input & \scriptsize \makecell{Implicit Disparity\\ \Cref{algo:layered_disp}} & \scriptsize Input & \scriptsize \makecell{Implicit Disparity\\ \Cref{algo:layered_disp_alt}} \\
        \includegraphics[width=.22\textwidth,height=.17\textwidth,trim={0 67.8cm 0 0},clip,valign=c]{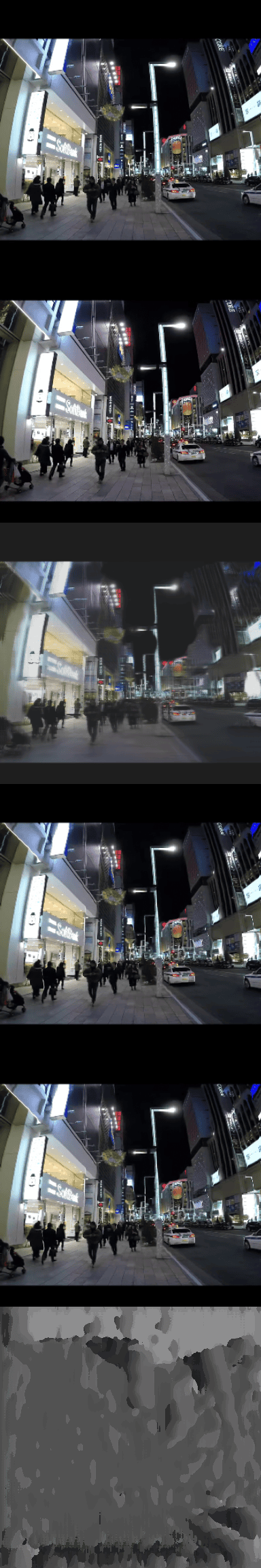}
        &
        \includegraphics[width=.22\textwidth,height=.17\textwidth,trim={0 0 0 67.8cm},clip,valign=c]{misc/implicit_disp_compare/v5_new_mask_35200.png}
        &
        \includegraphics[width=.22\textwidth,height=.17\textwidth,trim={0 67.8cm 0 0},clip,valign=c]{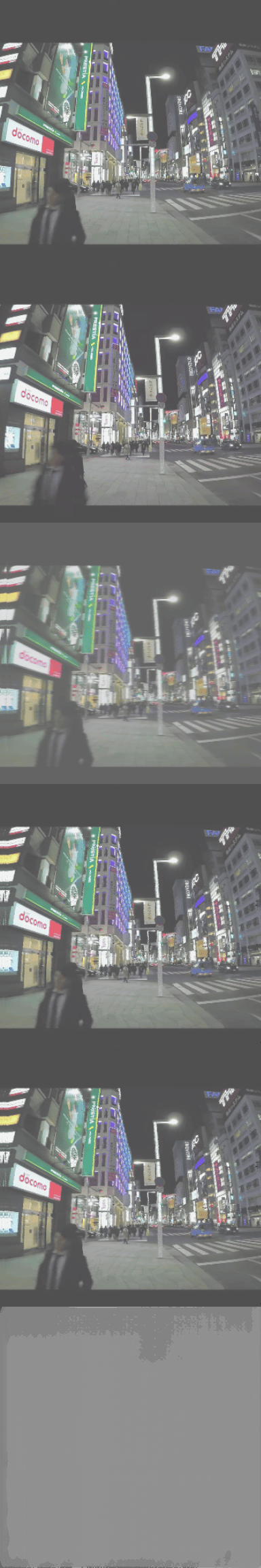}
        &
        \includegraphics[width=.22\textwidth,height=.17\textwidth,trim={0 0 0 67.8cm},clip,valign=c]{misc/implicit_disp_compare/v5_old_mask_35200.png}\\
        \includegraphics[width=.22\textwidth,height=.17\textwidth,trim={0 67.8cm 0 0},clip,valign=c]{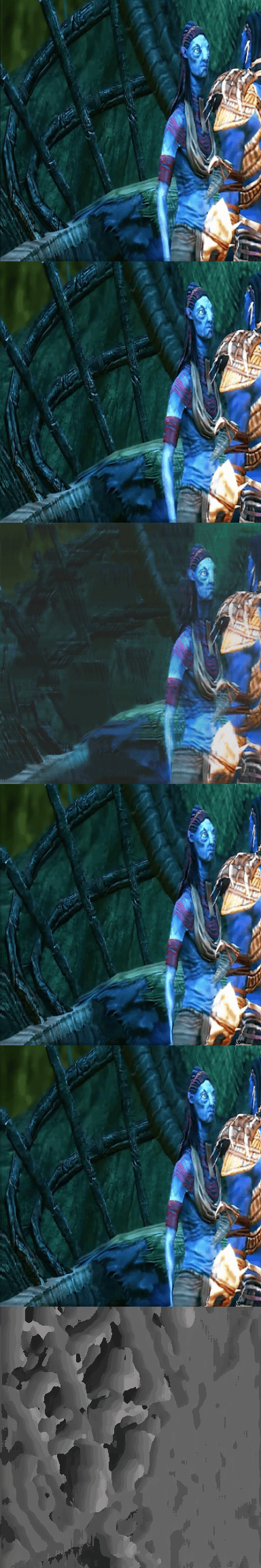}
        &
        \includegraphics[width=.22\textwidth,height=.17\textwidth,trim={0 0 0 67.8cm},clip,valign=c]{misc/implicit_disp_compare/v5_new_mask_36000.png}
        &
        \includegraphics[width=.22\textwidth,height=.17\textwidth,trim={0 67.8cm 0 0},clip,valign=c]{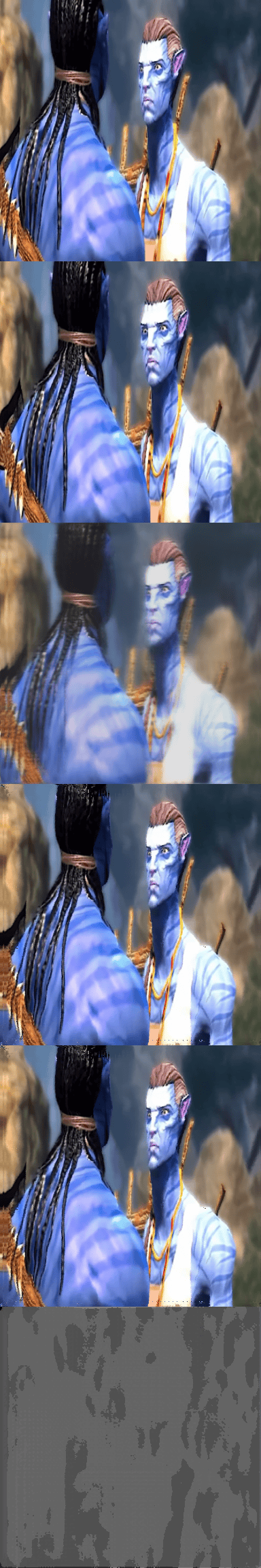}
        &
        \includegraphics[width=.22\textwidth,height=.17\textwidth,trim={0 0 0 67.8cm},clip,valign=c]{misc/implicit_disp_compare/v5_old_mask_36000.png}\\
        \includegraphics[width=.22\textwidth,height=.17\textwidth,trim={0 67.8cm 0 0},clip,valign=c]{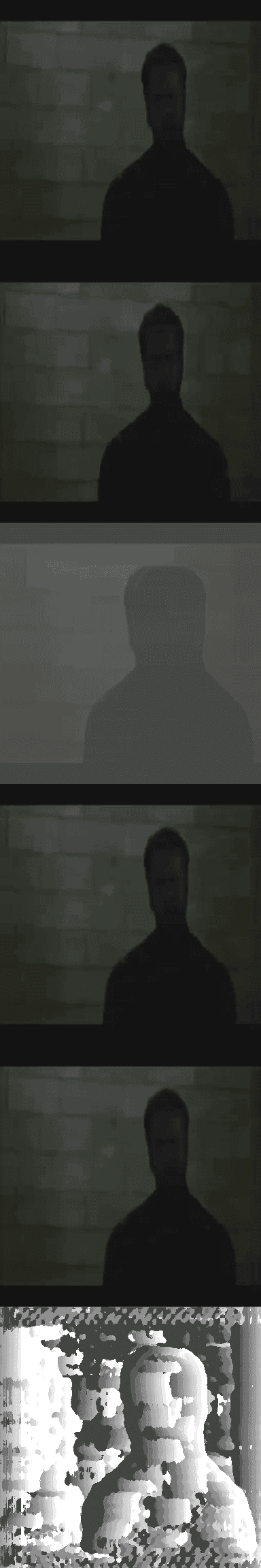}
        &
        \includegraphics[width=.22\textwidth,height=.17\textwidth,trim={0 0 0 67.8cm},clip,valign=c]{misc/implicit_disp_compare/v5_new_mask_35600.png}
        &
        \includegraphics[width=.22\textwidth,height=.17\textwidth,trim={0 67.8cm 0 0},clip,valign=c]{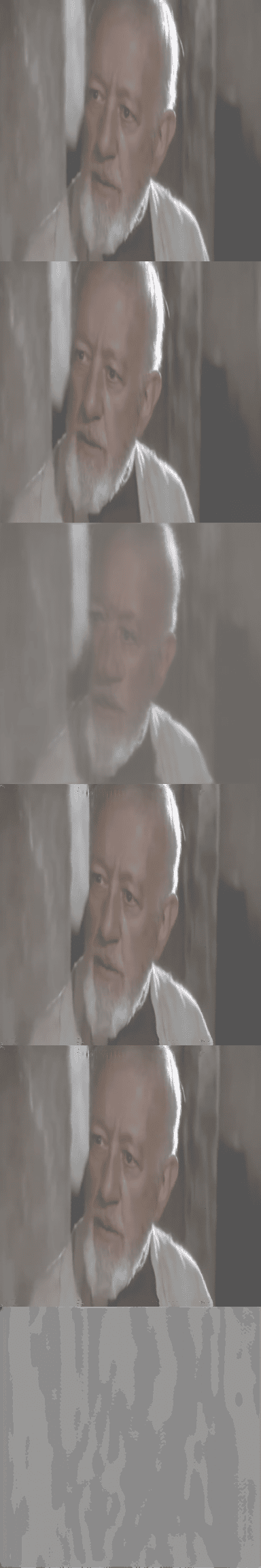}
        &
        \includegraphics[width=.22\textwidth,height=.17\textwidth,trim={0 0 0 67.8cm},clip,valign=c]{misc/implicit_disp_compare/v5_old_mask_37600.png}\\
    \end{tblr}
        \caption{Visual demonstration of the implicit disparity output for different masking strategies. The implicit disparity map contains multiple channels and we apply $argmax$ to obtain the visual output.}
        \label{fig:context_disp_diff}
\end{figure}

To avoid having multiple pixels being mapped to the same pixel location $i,j$, we use an algorithm to produce $[0, 1]$ masks so that different layers cannot interfere with each other as shown in~\Cref{algo:layered_disp}.
In addition, we further tested another design where the mask value selection algorithm~\Cref{algo:layered_disp_alt} generates mask values $\in \{-1, 0, 1\}$, to allow more interactions between layers.
However, though~\Cref{algo:layered_disp_alt} can better resolve complicated scenarios, we found the intermediate implicit disparity layers often fail to resolve disparities correctly, as shown in~\Cref{fig:context_disp_diff}.
In general, we found that the~\Cref{algo:layered_disp_alt} tries to weaken the disparity cues, resulting in smoother output with weaker or wrong disparity maps.

\begin{algorithm}[h]
\SetAlgoLined
    \PyComment{number\textunderscore layered\textunderscore disparity: the number of disparity layers.} \\
    \PyComment{warped\textunderscore output: `BDTCHW`. A stack of images warped by layered disparities. D is the number of disparity layers.} \\
    \PyComment{warped\textunderscore mask: `BDTCHW`. A stack of masks warped by layered disparities. D is the number of disparity layers.} \\
    \PyCode{layered\textunderscore mask = zeros\textunderscore like(output\textunderscore mask)} \\
    \PyCode{total\textunderscore mask = zeros\textunderscore like(output\textunderscore mask)} \\
    \PyCode{for i in range(number\textunderscore layered\textunderscore disparity):} \\
    \Indp   % start indent
        \PyCode{if i == 0:} \\
        \Indp   % start indent
            \PyCode{layered\textunderscore mask[:, i] = warped\textunderscore mask[:, i]} \\
            \PyCode{total\textunderscore mask[:, i] = warped\textunderscore mask[:, i]} \\
        \Indm % end indent, must end with this, else all the below text will be indented
        \PyCode{else:} \\
        \Indp   % start indent
            \PyCode{total\textunderscore mask[:, i] = logical\textunderscore or(warped\textunderscore mask[:, i], layered\textunderscore mask[:, i - 1])} \\
            \PyCodeHighlight{layered\textunderscore mask[:, :, i] = total\textunderscore mask[:, :, i] - output\textunderscore mask[:, :, i - 1]} \\
        \Indm % end indent, must end with this, else all the below text will be indented
    \Indm % end indent, must end with this, else all the below text will be indented
    \PyCode{output = layered\textunderscore mask * warped\textunderscore output} \\
\caption{Synthesis from layered disparities.}
\label{algo:layered_disp_alt}
\end{algorithm}

\subsection{Context Fusion Module}

\begin{figure}[h]
     \centering

     \begin{tblr}{colspec={Q[c]Q[c]Q[c]}, colsep = {.3pt}, rowsep = {.3pt}}
        GT Left & w Context Fusion & w/o Context Fusion \\
        \includegraphics[width=.25\textwidth,height=.20\textwidth,trim={0 67.8cm 0 0},clip,valign=c]{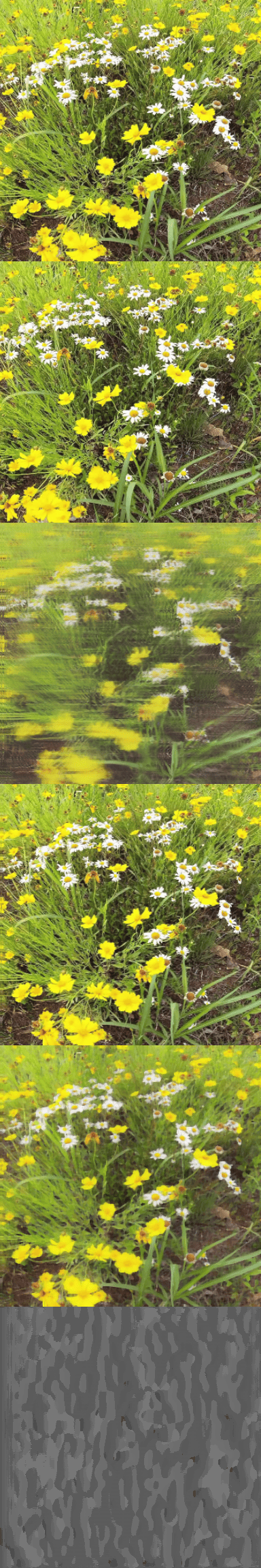}
        &
        \includegraphics[width=.25\textwidth,height=.20\textwidth,trim={0 13.6cm 0 54.2cm},clip,valign=c]{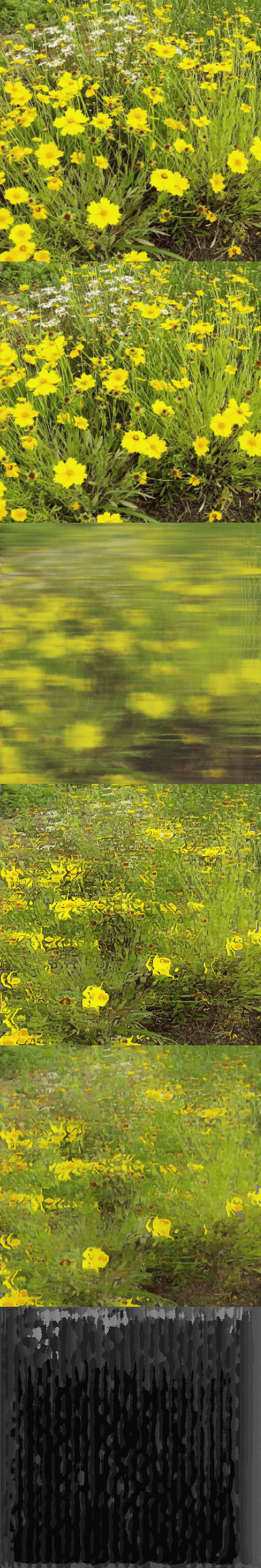}
        &
        \includegraphics[width=.25\textwidth,height=.20\textwidth,trim={0 13.6cm 0 54.2cm},clip,valign=c]{misc/good_stereo_step20600.png}\\
    \end{tblr}
        \caption{Visual demonstration of the failed edge cases of context fusion.}
        \label{fig:failed_algo}
\end{figure}

As shown in~\Cref{tab:main_table}, the inclusion of the context fusion module significantly enhances the overall statistical performance. Moreover, as demonstrated in the accompanying videos, this module greatly improves the temporal consistency of the generated videos. However, we observed potential artifacts in frames with complex feature patterns, as illustrated in~\Cref{fig:failed_algo}. We suspect that these edge cases could be mitigated with a larger training dataset that includes greater internal variance, allowing the model to better handle such intricate scenarios.

% \begin{table}[h]
%     \centering
%     \small
%     \begin{tabular}{l|c c c c}
%         \toprule
%         & L1 $\downarrow$ & SSIM $\uparrow$ & PSNR $\uparrow$\\
%         \midrule
%          Algorithm 1 & 0.0588 & 0.5959 & 21.6649 \\
%          Algorithm 2 & \cellcolor{green!25} 0.0581 & \cellcolor{green!25} 0.6007 & \cellcolor{green!25} 21.7725 \\
%         \bottomrule
%     \end{tabular}
%     \caption{Benchmark results. The best results are highlighted in green.}
%     \label{tab:algo_tab}
% \end{table}

\subsection{Flow-Guided Feature Propagation}

% \subsection{Flow-Guided Feature Propagation}
% \label{sec:flow}

Video feature propagation and deformation have shown their effectiveness for many video-based tasks~\cite{xue2019video,wang2019edvr,haris2019recurrent}.
The flow-guided deformation concept is particularly suitable for the stereo conversion scenario as the pixel shifting nature according to the disparities.
Similar to E$^2$FGVI~\cite{liCvpr22vInpainting} and ProPainter~\cite{zhou2023propainter}, we use a similar design of flow-guided feature propagation module, that features bi-directional optical flow-guided deformable alignments that built on top of deformable convolution networks (DCN)~\cite{dai2017deformable,zhu2019deformable}.

Given extracted features $\{E_t|t=1...T\}$ from a feature encoder where $T$ is the total number of frames.
Under the context of stereo conversion, the forward flow $F_{t\rightarrow t+1}$ helps to track the movement of occluded regions from frame $t$ to frame $t+1$. When the pixels within the occluded areas of frame $t$ are found in the valid regions of frame $t+1$, this information can be utilized effectively by warping the backward propagation feature $\hat{E}_b^{t+1}$ from frame $t+1$ back to frame $t$, guided by the forward flow ${F}_{t\rightarrow t+1}$.
On top of E$^2$FGVI's approach, we include flow validation maps $M_{t+1\rightarrow t}$ by consistency check introduced by ProPainter.
The consistency check compares the forward and backward optical flows to ensure the correctness of the used optical flows. Similar to~\Cref{eq:consistency}, the consistency error is computed as follows:
\begin{equation}
    \mathcal{E}_{t\rightarrow t+1}(p) = ||F_{t\rightarrow t+1}(p)) + F_{t+1\rightarrow t}(p+F_{t\rightarrow t+1}(p))||_2^2,
\end{equation}
where $p$ is pixel positions of the frame.
Then the flow deformation offsets $\Tilde{o}_{t\rightarrow t+1})$ are computed with the DCN network, where a concatenation of the forward flow ${F}_{t\rightarrow t+1}$, backward propagation feature $\hat{E}_b^{t+1}$, warped backward feature $\mathcal{W}(\hat{E}_b^{t+1}, {F}_{t\rightarrow t+1})$, and flow validation maps $M_{t+1\rightarrow t}$ is used as the condition, where $\mathcal{W}$ is warping operation. The flow-guided alignment propagation is then:
\begin{equation}
    \hat{E}_b^{t} = \mathcal{R}(\mathcal{D}(\hat{E}_b^{t+1};F_{t\rightarrow t+1} + \Tilde{o}_{t\rightarrow t+1}),f_t),
\end{equation}
where $\mathcal{D}(\cdot)$ is the deformable convolution layers and $\mathcal{R}(\cdot)$ fuses the aligned and current features.
% Unlike inpainting methods such as ProPainter, the inpainting masks are not used as a condition in the flow guidance module.

However, we found the disparity cannot be learned with those flow-guided propagation modules. We suspect the feature map deformation and alignment can break the internal disparity features, resulting in a failed learning of the implicit disparity maps.

\subsection{Different Backbones For the Depth Branch}

We provide additional results in~\cref{tab:suppl_table}. We experimented with MiDaS instead of DepthAnything. The results indicate that different depth estimation backbones do not affect the performance of our proposed method.

\begin{table}[h]
    \centering
    \small
    \begin{tabular}{l|c| c c c}
        \toprule
        & Depth Backbone & L1 $\downarrow$ & SSIM $\uparrow$ & PSNR $\uparrow$\\
        \midrule
         Ours w/o context fusion & MiDaS & 0.0590 & \cellcolor{green!25} 0.6014 & 21.6572 \\
         Ours w/o context fusion & DepthAnything & \cellcolor{green!25} 0.0588 &  0.5959 & \cellcolor{green!25} 21.6649 \\
        \bottomrule
    \end{tabular}
    \caption{Additional results. The best results are highlighted in green.}
    \label{tab:suppl_table}
\end{table}

\end{document}